\documentclass[preprint,review,12pt]{elsarticle}

\usepackage{amssymb}

\usepackage{amsthm}
\usepackage{pifont}
\usepackage{subfigure}
\usepackage{bbm}
\usepackage{bbding}
\usepackage{utfsym}
\usepackage{hyperref}
\usepackage{longtable}
\usepackage{pdflscape}
\usepackage{rotating}
\usepackage{tabularx} 
\usepackage{amssymb}
\usepackage{makecell}

\usepackage{algorithm}  
\usepackage{algorithmic}

\makeatletter
\newenvironment{breakablealgorithm}
{
		\begin{center}
			\refstepcounter{algorithm}
			\hrule height.8pt depth0pt \kern2pt
			\renewcommand{\caption}[2][\relax]{
				{\raggedright\textbf{\ALG@name~\thealgorithm} ##2\par}
				\ifx\relax##1\relax 
				\addcontentsline{loa}{algorithm}{\protect\numberline{\thealgorithm}##2}
				\else 
				\addcontentsline{loa}{algorithm}{\protect\numberline{\thealgorithm}##1}
				\fi
				\kern2pt\hrule\kern2pt
			}
		}{
		\kern2pt\hrule\relax
	\end{center}
}
\makeatother

\usepackage{xcolor}
\usepackage{array}
\usepackage{siunitx}
\usepackage{numprint}
\usepackage{booktabs}

\usepackage{csquotes}
\usepackage{graphicx}
\usepackage{amsmath}
\usepackage{multicol}
\usepackage{multirow}
\usepackage{csquotes}

\usepackage[accsupp]{axessibility} 

\usepackage{amsmath}
\journal{Engineering Applications of Artificial Intelligence}

\begin{document}

\begin{frontmatter}

\title{Boundary-Refined Prototype Generation: A General End-to-End Paradigm for Semi-Supervised Semantic Segmentation}

\author[a]{Junhao Dong}\author[a]{Zhu Meng}\author[a]{Delong Liu}\author[a]{Jiaxuan Liu}\author[a,b]{Zhicheng Zhao\corref{cor1}}\author[a,b]{Fei Su}

\affiliation[a]{organization={School of Artificial Intelligence},
            addressline={Beijing University of Posts and Telecommunications}, 
            city={Beijing},
            postcode={100876}, 
            country={China}}

\affiliation[b]{organization={Beijing Key Laboratory of Network System and Network Culture},
            city={Beijing},
            country={China}}
\cortext[cor1]{Corresponding author.}
            
\begin{abstract}

Semi-supervised semantic segmentation has attracted increasing attention in computer vision, aiming to leverage unlabeled data through latent supervision. To achieve this goal, prototype-based classification has been introduced and achieved lots of success. However, the current approaches isolate prototype generation from the main training framework, presenting a non-end-to-end workflow. Furthermore, most methods directly perform the K-Means clustering on features to generate prototypes, resulting in their proximity to category semantic centers, while overlooking the clear delineation of class boundaries. To address the above problems, we propose a novel end-to-end boundary-refined prototype generation (BRPG) method. Specifically, we perform online clustering on sampled features to incorporate the prototype generation into the whole training framework. In addition, to enhance the classification boundaries, we sample and cluster high- and low-confidence features separately based on confidence estimation, facilitating the generation of prototypes closer to the class boundaries. Moreover, an adaptive prototype optimization strategy is proposed to increase the number of prototypes for categories with scattered feature distributions, which further refines the class boundaries. Extensive experiments demonstrate the remarkable robustness and scalability of our method across diverse datasets, segmentation networks, and semi-supervised frameworks, outperforming the state-of-the-art approaches on three benchmark datasets: PASCAL VOC 2012, Cityscapes and MS COCO. The code is available at \url{https://github.com/djh-dzxw/BRPG}.
\end{abstract}

\begin{graphicalabstract}

\end{graphicalabstract}

\begin{highlights}
\item Research highlight 1
\item Research highlight 2
\end{highlights}


\end{frontmatter}

\section{Introduction}
Semantic segmentation is an essential vision task. In recent years, along with advances in deep learning, rapid progress has been made in segmentation methods \cite{Deeplab,fcn,unet,pyramid}; nevertheless, it requires large-scale dense pixel-level annotations, which are time-consuming and costly to obtain. To alleviate this problem, much research effort \cite{ccvc,cps,density,semi1,semi2,semi3,u2pl} has been dedicated to semi-supervised manner, which aims to develop effective training pipelines with limited annotations and numerous unlabeled data.

Early semi-supervised approaches mainly follow the scheme of adversarial learning \cite{gan1,gan2,gan3}, where a discriminator is introduced to distinguish the segmentation maps (generated by the segmentation network) from the real ones (i.e., ground truth), thereby enhancing the prediction accuracy. Recently, a slice of methods have focused on consistency regularization \cite{semi3,cct,ps-mt} and pseudo-labeling \cite{pseudoseg,eln,ecn,st++}. The former involves applying perturbations at different levels (e.g., input, feature and network) to enforce the consistency of predictions, while the latter produces pseudo-labels for unlabeled images to supplement training samples. Both methods learn from unlabeled data in a pseudo-supervision manner.

Additionally, many methods have introduced pixel-level contrastive learning \cite{semi1,u2pl,contextaware,contrast2,reco} as an auxiliary task to augment supervision. Although the optimization of interclass separability can be achieved by contrastive learning, the issue of intraclass compactness is ignored, resulting in a scattered distribution of features within each class. Therefore, \cite{prototype} proposes a prototype-based solution, where each class is abstracted by a set of high-dimensional features, i.e., prototypes, to describe rich intraclass semantics, thereby facilitating the learning of compact representations. 

The representative prototype-based approach \cite{pcr} usually includes two stages: prototype initialization, and model retraining. Prototype initialization can be divided into three steps: (1) the training of a supervised segmentation network, (2) new features, i.e., before they are fed into the classification head, are extracted on the trained model, (3) $K$ initial prototypes are created by the K-Means algorithm after random sampling the features. In the second stage, a semi-supervised model is retrained, benefiting from prototype-based classification with the initialized prototypes that are continuously updated throughout the training process.

However, this method prompts two questions: \ding{172} Is it necessary to train a separate feature extractor for prototype initialization? \ding{173} Is there any superior method to generate prototypes compared to random sampling and direct K-Means clustering?

For question \ding{172}, \cite{pcr} uses a dedicated feature extractor to isolate the prototype initialization process from the semi-supervised training framework and presents a non-end-to-end workflow. In contrast, in this work, we propose an end-to-end prototype generation method that can be integrated into the whole process of semi-supervised training, and achieve comparable performance. Specifically, without utilizing the feature representations before the classification head, an additional feature extraction subnet (a.k.a, feature head) is introduced to the segmentation model (Section \ref{sec:ablation study} demonstrates the superiority of this configuration). During the sampling stage, pixel-level features from both labeled and unlabeled data are simultaneously sampled to ensure feature alignment. We employ a category-wise memory bank to store these features, facilitating the generation of class prototypes with online K-Means clustering for subsequent training. Note that the feature head is not trained before the sampling stage. Thus, the sampled features solely depend on the shared encoder and the well-designed structure of this subnet. Surprisingly, the proposed approach achieves a comparable performance with \cite{pcr} (76.77 vs. 77.16 on the 366 split of the PASCAL VOC 2012 Dataset \cite{pascal}). This demonstrates that the randomly selected features during training can sufficiently capture the distribution of each category, obviating the need for a separately trained feature extractor.

\begin{figure*}[t]
\centering
\subfigure[Random]{\includegraphics[width=0.3\paperwidth]{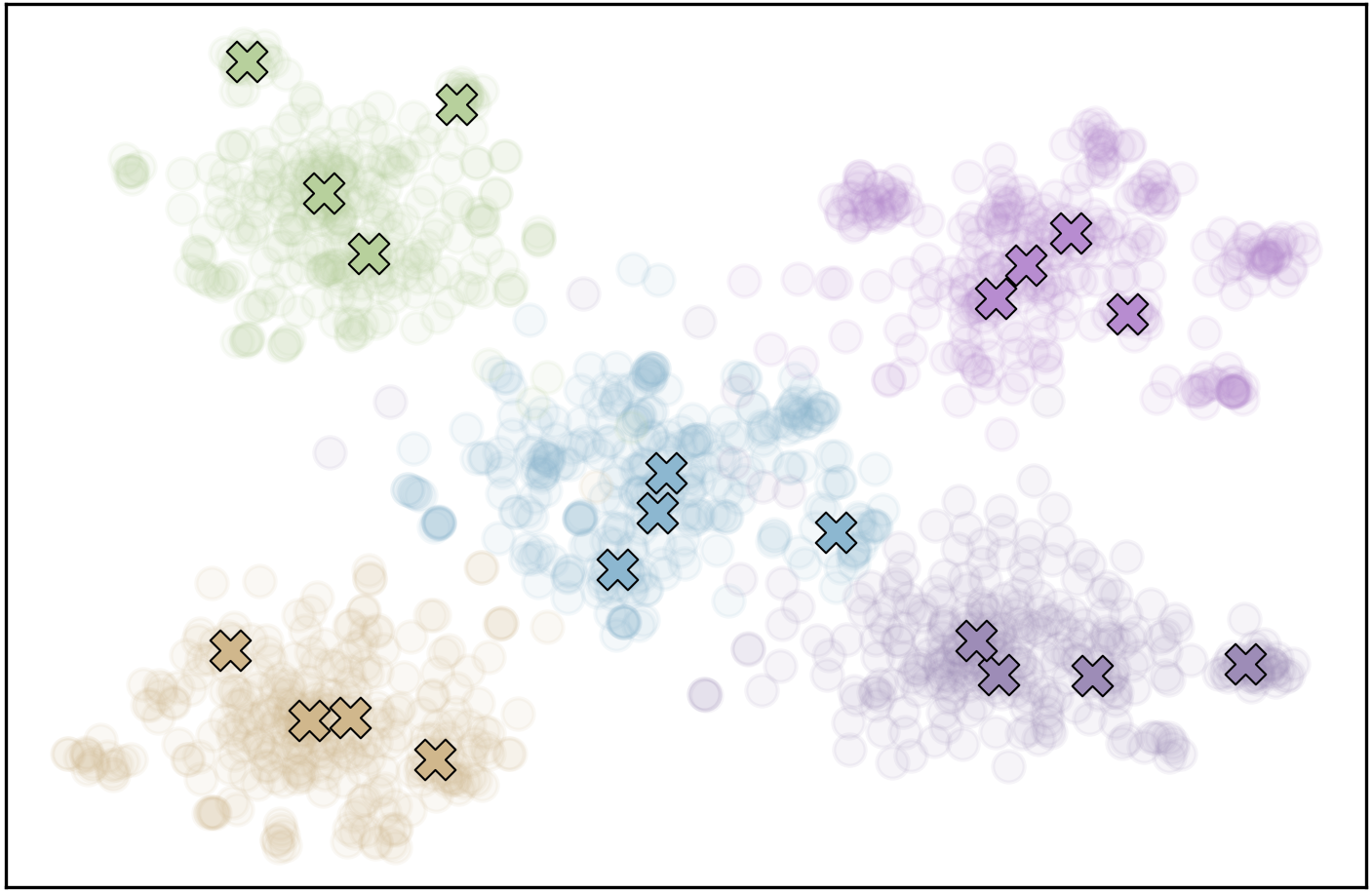}}
\hspace{5mm}
\subfigure[Confidence-based]
{\includegraphics[width=0.3\paperwidth]{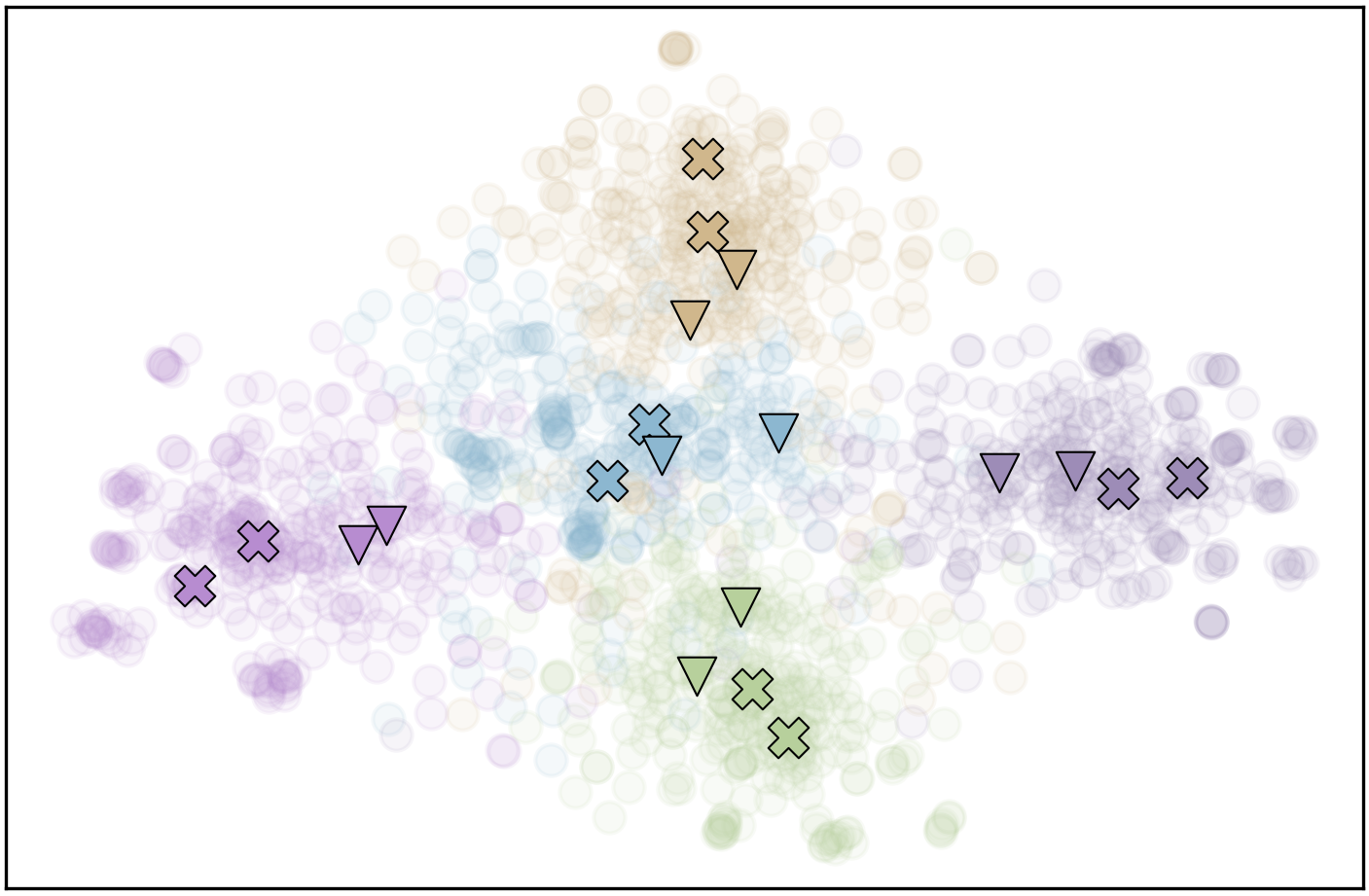}}\\
\caption{Visualization of the feature embeddings sampled by the two manners. (a) Random sampling and clustering. (b) Separate sampling and clustering based on a confidence threshold set as 0.8. The symbol $``\text{$\boldsymbol{\times}$}"$ represents the generated prototypes in (a) and the high-confidence prototypes in (b), while $``\triangledown"$ denotes the low-confidence prototypes, typically closer to the classification boundaries.}
\label{fig1}
\end{figure*}

\begin{figure*}[!t]
\centering
\includegraphics[width=0.6\paperwidth]{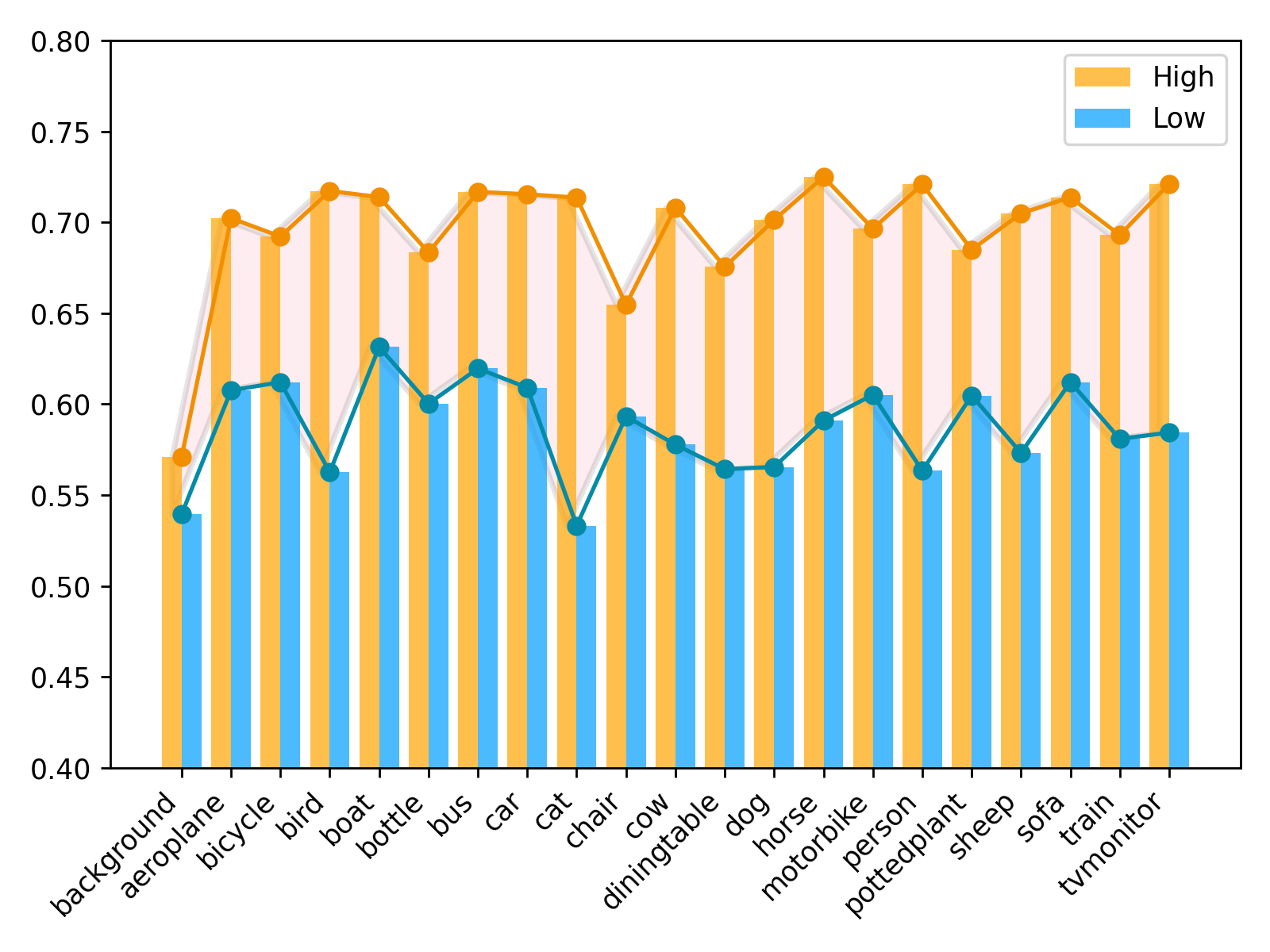}
\caption{Mean cosine similarities between high- and low-confidence features and the class centers on PASCAL VOC 2012. Lower values indicate that features tend to deviate further from the class centers.}
\label{fig2}
\end{figure*}

For question \ding{173}, considering that prototype-based learning is essentially a nonparametric classification based on the similarity between features and non-learnable prototypes; consequently, a superior initialization of prototypes can lead to improved classification performance. Although the direct use of K-Means has considered intraclass variance, the generated prototypes tend to be close to category semantic centers, overlooking the clear delineation of class boundaries (as shown in Fig.~\ref{fig1} (a)). Moreover, according to \cite{badgan}, leveraging low-quality samples in low-density regions can help refine the model's class boundaries, resulting in an improvement of classification performance. Inspired by this, to enhance the classification boundaries, we propose to specially sample low-confidence features, facilitating the generation of prototypes closer to the class boundaries. Specifically, the features of each class are divided into two groups via confidence computation to perform sampling and clustering. As shown in Fig.~\ref{fig1}, compared to random sampling and clustering, our method effectively makes the low-confidence prototypes move closer to the class boundaries, capturing more difficult samples. The statistical graph in Fig.~\ref{fig2} also confirms that low-confidence features tend to deviate further from the category center. The improved performance of 
the confidence-based prototype generation in Section~\ref{sec:ablation study} validates its superiority. In addition, to further refine the class boundaries, an adaptive prototype optimization scheme is proposed to increase the number of generated prototypes for categories with scattered distributions.

Our approach is implemented based on a popular semi-supervised framework, Mean Teacher \cite{meanteacher}, using the DeepLabV3+ \cite{deeplabv3+} segmentation network. It demonstrates remarkable performance across three benchmark datasets (PASCAL VOC 2012 \cite{pascal}, Cityscapes \cite{cityscapes}, and MS COCO \cite{coco}) of different scales, scenes, and complexities, consistently outperforming the state-of-the-art methods. Moreover, utilizing the transformer-based SegFormer \cite{segformer} as the segmentation network can yield similarly impressive or even better results, validating its robustness across various segmentation architectures. Notably, the proposed method is also highly scalable and can be easily incorporated into diverse semi-supervised frameworks. The successful integration with the classical frameworks FixMatch \cite{fixmatch} and CPS \cite{cps} serves as compelling examples, resulting in significant enhancements over the baseline models. Our contributions can be summarized as follows:

\begin{enumerate}
\item{We propose an end-to-end prototype generation method that can be integrated into the whole training framework. Furthermore, a comprehensive end-to-end pipeline is presented for semi-supervised semantic segmentation, including the process of pretraining, prototype generation, and prototype-based learning. With these simple yet effective strategies, the proposed method achieves a comparable performance with the current non-end-to-end prototype-based methods.}
\item{From the classification essence of prototype learning, we present a novel perspective on optimizing this process, i.e., prototype generation. To enhance classification boundaries, we propose to leverage low-quality samples for generating prototypes close to the class boundaries, thereby improving classification performance.}
\item{Two simple yet effective strategies, confidence-based prototype generation and adaptive prototype optimization, are proposed to refine the classification boundaries, leading to enhanced segmentation results.}
\item{Our final approach consistently outperforms the state-of-the-art methods on three benchmark datasets: PASCAL VOC 2012 \cite{pascal}, Cityscapes \cite{cityscapes}, and MS COCO \cite{coco}. Moreover, extensive experiments conducted on different segmentation networks and semi-supervised frameworks illustrate the scalability and applicability of our approach.}
\end{enumerate}

\section{Related Work}
\subsection{Semi-Supervised Learning}
Semi-supervised learning (SSL) \cite{semilearn} has been widely applied to scenarios where labeled data are limited but a large number of unlabeled data can be acquired. The key is to effectively utilize the latent supervision in unlabeled data. Existing methods primarily belong to two representative families: consistency regularization \cite{meanteacher,temporal,mixmatch,ict,uda,vat} and entropy minimization \cite{fixmatch,lee2013pseudo,memory,curriculum}. Consistency regularization methods encourage the network to produce consistent predictions on the same example under different augmentations. For example, MixMatch \cite{mixmatch} and VAT \cite{vat} apply perturbations at the input and feature levels individually to regularize the outputs of the network. The mean teacher \cite{meanteacher} incorporates an additional weight-averaged network to regularize the prediction consistency. Entropy minimization focuses on the implementation of self-training with generated pseudo-labels. Typically, FixMatch \cite{fixmatch} leverages pseudo labels generated from weakly augmented unlabeled images to supervise their strongly augmented versions for robust learning. Moreover, graph-based regularization \cite{gcn,iscen2019label}, deep generative models \cite{badgan,dgm,adgm,lobalgan} and self-supervised learning \cite{s4l,simclr,simclrv2} are also important branches of semi-supervised learning. Our method builds upon the training frameworks of Mean Teacher \cite{meanteacher} and FixMatch \cite{fixmatch}, illustrating their continued effectiveness.

\subsection{Semi-Supervised Semantic Segmentation}
The success of SSL can be easily expanded to semantic segmentation with the principles of consistency regularization and pseudo-labeling. PS-MT \cite{ps-mt} establishes consistency between two teacher models and a student model and employs a challenging combination of input data, features and network perturbations to greatly improve generalization. CPS \cite{cps} imposes consistency on two segmentation networks perturbed with different initializations through cross pseudo supervision. Instead of introducing an extra network, CCVC \cite{ccvc} relies on a two-branch co-training network to extract features from irrelevant views and enforces consistent predictions through cross-supervision. UniMatch \cite{unimatch} proposes a consistency framework that divides image and feature perturbations into two independent streams to expand the perturbation space. It further presents a dual-stream strong perturbation to fully explore the predefined image-level augmentations.
However, due to the density and complexity of semantic segmentation tasks, the generated pseudo-labels inevitably contain a large amount of noise, severely limiting the semi-supervised training. A prevalent scheme is to filter out unreliable pseudo-labels by setting the confidence \cite{pcr} or entropy threshold \cite{u2pl}. In addition, some recent studies \cite{eln,ecn,gct} have proposed using an additional error detection network to refine the generated pseudo-labels. Nevertheless, most segmentation datasets \cite{pascal,cityscapes,coco} suffer from a problem of long-tailed class distribution, characterized by extreme pixel-wise class imbalance in training samples. Consequently, deep models trained on such data tend to produce reliable pseudo-labels biased to majority classes, further exacerbating the class-bias problem. To address this issue, \cite{dars} leverages prior knowledge of the dataset and proposes aligning the class distribution of pseudo-labels with the true distribution through class-wise thresholding and random sampling. \cite{ael} designed a series of adaptive data augmentations, such as adaptive CutMix and Copy-Paste, along with a balanced sampling strategy, to enhance the supervision for underperforming categories. \cite{usrn} presents cluster balanced subclass distributions to train a class-unbiased segmentation model.

Recently, inspired by contrastive learning, a number of studies have focused on regularizing feature representations in the latent space. ReCo \cite{reco} enhances pixel-level contrastive learning by considering semantic relationships between different classes, effectively alleviating the uncertainty among confused classes. $\text{U}^2\text{PL}$ \cite{u2pl}, on the other hand, incorporates unreliable pixels into the storage queue of negative samples to make full use of unlabeled data. In addition, \cite{pcrl} first proposes a probabilistic representation contrastive learning framework, aiming to mitigate the risk of unreliable pseudo-labels through Gaussian modeling. However, regularized representations only cover contextual or batch-level semantics of different classes and lack category-level semantics across the whole dataset. Consequently, the optimization is only implemented on interclass relationships, overlooking intraclass compactness.

Prototype learning provides an elegant resolution to this issue. Recently, it has been integrated into semantic segmentation. In particular, \cite{prototype} first proposes to represent each category with a set of non-learnable class prototypes and make dense predictions via nearest prototype retrieval. PCR \cite{pcr} applies this scheme to semi-supervised semantic segmentation and achieves remarkable performance. However, it employs a non-end-to-end workflow for prototype generation and overlooks the clear delineation of class boundaries. In contrast, we propose a novel way to generate prototypes with refined class boundaries and incorporate all steps into a unified training process, leading to significant improvements in performance.

\begin{figure*}[!t]
\centering
\includegraphics[width=0.6\paperwidth]{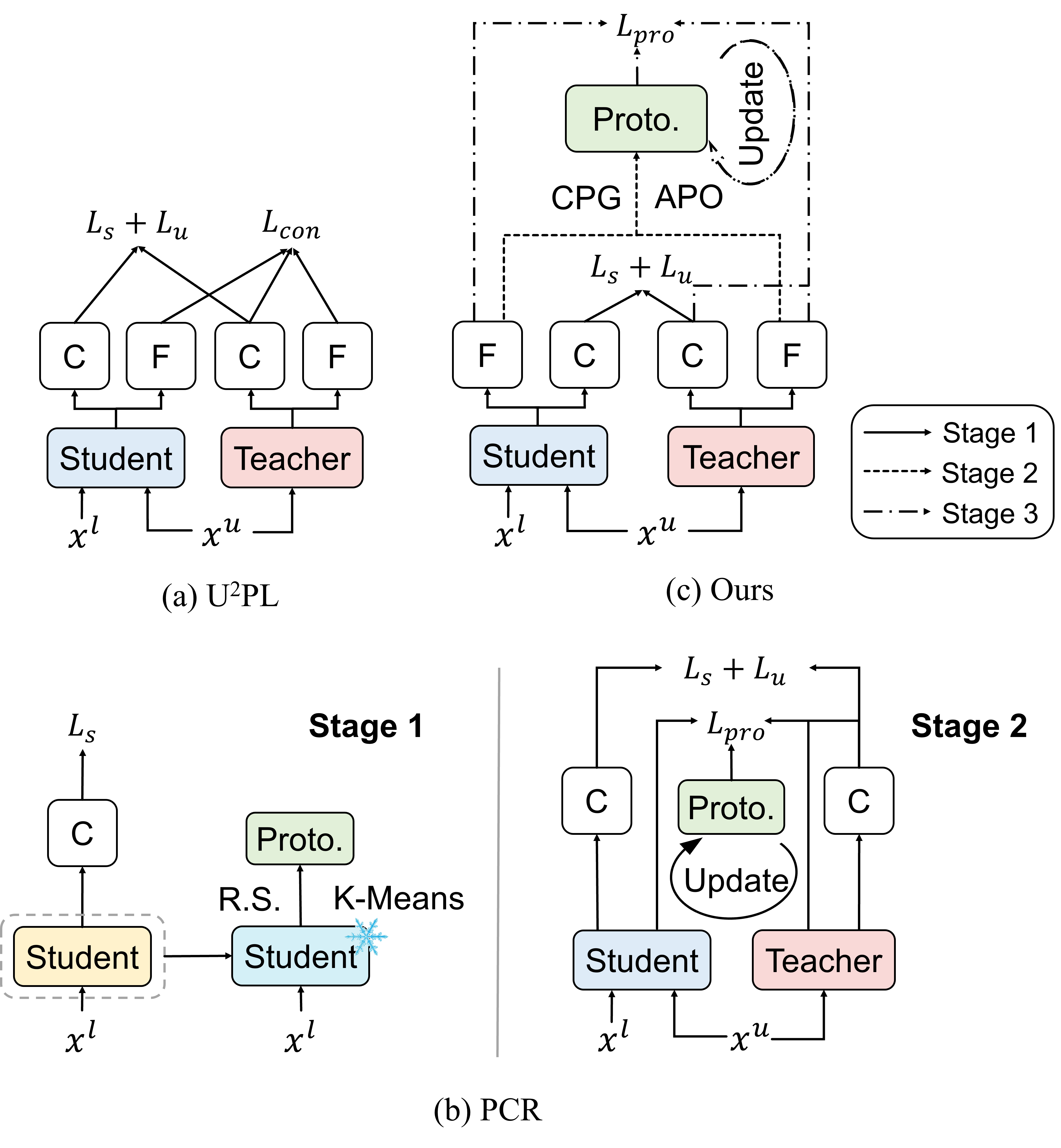}
\caption{Concise flowcharts of (a) $\text{U}^2\text{PL}$ \cite{u2pl}, (b) PCR \cite{pcr}, and (c) Ours. Student: Student model's encoder. Teacher: Teacher model's encoder. C: Classification head. F: Feature head. Proto.: Class prototypes. R.S.: Random sampling.}
\label{fig-flowchart}
\end{figure*}

\section{Method}\label{sec:Method}

\subsection{Preliminary}\label{sec:Preliminaries}
\subsubsection{Prior Works}
In this section, we discuss the relationships between the proposed method and the most relevant works, $\text{U}^2\text{PL}$ \cite{u2pl} and PCR \cite{pcr}, used in its design.

\textbf{Compared with $\text{U}^2\text{PL}$ \cite{u2pl}.} Apart from the inherent classification head, $\text{U}^2\text{PL}$ incorporates an additional feature head into the segmentation network. Furthermore, it is also implemented based on the semi-supervised framework of Mean Teacher \cite{meanteacher}, as shown in Fig.~\ref{fig-flowchart} (a). During the training stage, pixel-level features extracted by the feature head are employed for contrastive learning. This process involves pulling the sampled instances closer to their positive samples within the same category while pushing them away from the negative samples of different categories. Nevertheless, the pixel-wise contrastive learning is not utilized because its positive samples are only composed of batch-level category centers, lacking the broader category-level semantics across the entire dataset. This results in only enhancing interclass separability, while the feature distribution within each class still lacks compactness. In our work, prototype learning is employed to optimize the intraclass compactness. Specifically, the feature head and a memory bank are utilized to sample and store features within a specific epoch range. After the sampling stage, online K-Means clustering is implemented to generate multiple prototypes for each category, capturing rich semantics. The subsequent prototype-based contrastive learning facilitates the clustering of features from the same category around their respective prototypes, thereby enhancing intraclass compactness. Experimental results also confirm the superior performance of our approach compared to $\text{U}^2\text{PL}$.

\textbf{Compared with PCR \cite{pcr}.} Both our method and PCR are prototype-based semantic segmentation approaches. However, PCR separates prototype generation from the training framework, demonstrating a non-end-to-end workflow, as shown in Fig.~\ref{fig-flowchart} (b). In our proposed method, the process of prototype generation is integrated into the overall training framework, as shown in Fig.~\ref{fig-flowchart} (c). Instead of using features produced by the encoder, we conduct online sampling and storage of features with the same feature head as in the $\text{U}^2\text{PL}$ method. Furthermore, in PCR, the use of random sampling and direct K-Means clustering may result in the proximity of generated prototypes to category semantic centers, while overlooking the clear delineation of class boundaries. To address this issue, we propose a confidence-based prototype generation (CPG) method. Specifically, to enhance the classification boundaries, we perform separate sampling and clustering of high- and low-confidence features based on confidence estimation. This facilitates the generation of prototypes that are closer to the class boundaries. In addition, an adaptive prototype optimization (APO) strategy is introduced to increase the number of prototypes for categories exhibiting scattered feature distributions, which further refines the class boundaries. Subsequently, these generated prototypes are employed in prototype learning to optimize the feature head and the encoder. The improved experimental performance validates the effectiveness of the proposed method.

\subsubsection{Prototype-Based Learning}
Prototype-based learning can be regarded as a nonparametric classification process that allocates labels based on the similarity between sampled features and non-learnable prototypes. Assuming both features and prototypes are normalized representations, the probability of assigning sample $i$ to class $c$ can be formulated as:
\begin{equation}\label{eq:prototype essential}
p_i(c)=\frac{\exp \left(w_c^T f_i\right)}{\sum_{c^{\prime}=1}^C \exp \left(w_{c^{\prime}}^T f_i\right)},
\end{equation}
where $p_i(c)$ denotes the probability of the $i$-th pixel belonging to the $c$-th category, $f_i$ is the feature representation of pixel $i$, and $W=\left\{\left(w_c\right)\right\}_{c=1}^C$ represents the class prototypes in the feature space. Therefore, $w_c^T f_i$ measures the cosine distance (cosine similarity) between the feature $f_i$ and the class prototype $w_c$. Intriguingly, if consider $W$ as the learnable parameters of a classifier and $f_i$ as its input embedding, then the above formulation can precisely represent a parameterized softmax classifier.

It is worth noting that the prototype-based classifier and the parameterized classifier have distinct decision mechanisms. Specifically, the parameterized classifier allocates learnable parameters to different dimensions of the extracted feature representations, enabling it to focus more on discriminative dimensions, which may lead to overfitting to the specific feature dimensions, thereby hindering the generalization ability of the model. In contrast, the target of a prototype-based classifier is to make the generated features close to their respective prototypes, treating all dimensions equally. This approach may contribute to more robust features. Therefore, prototype learning can encourage the model to generate features from a different view, thereby enhancing the model's generalization performance.

Our proposed method also introduces an auxiliary branch (feature head) based on prototype learning. Furthermore, considering its classification essence, we introduce a novel boundary-refined prototype generation (BRPG) scheme to alleviate challenging samples in prototype-based classification and enhance the the model’s robustness.

\subsection{Overview}
In semi-supervised semantic segmentation, two sets of data are available for training a segmentation model: a manually labeled dataset $D_{L}=\left\{\left(x_{i}^{l}, y_{i}^{l}\right)\right\}_{i=1}^{L}$ and an unlabeled dataset ${D_{U}=\left\{x_{i}^{u}\right\}_{i=1}^{U}}$, where $L<<U$. To make full use of massive unlabeled data, the mean teacher \cite{meanteacher} framework is commonly used as the basic semi-supervised framework.

Our method is also based on mean teacher, and its concise process is illustrated in Fig.~\ref{fig-flowchart} (c). Specifically, it comprises three training stages. The first stage presents the pretraining process, where a combination of supervised learning and a pseudo-labeling strategy are utilized for semi-supervised training, which is consistently implemented in the following two stages. Meanwhile, the extra feature heads are introduced to extract pixel-level features from both labeled and unlabeled images. In the second stage, the extracted features are used to initialize class prototypes through two strategies: confidence-based prototype generation (CPG) and adaptive prototype optimization (APO). This constitutes the core of our method, and a comprehensive discussion is provided in Section \ref{sec:BRPG}. In the final stage, the generated prototypes are employed for prototype-based contrastive learning and continuously updated during training. 

\begin{figure*}[!t]
\centering
\includegraphics[width=0.63\paperwidth]{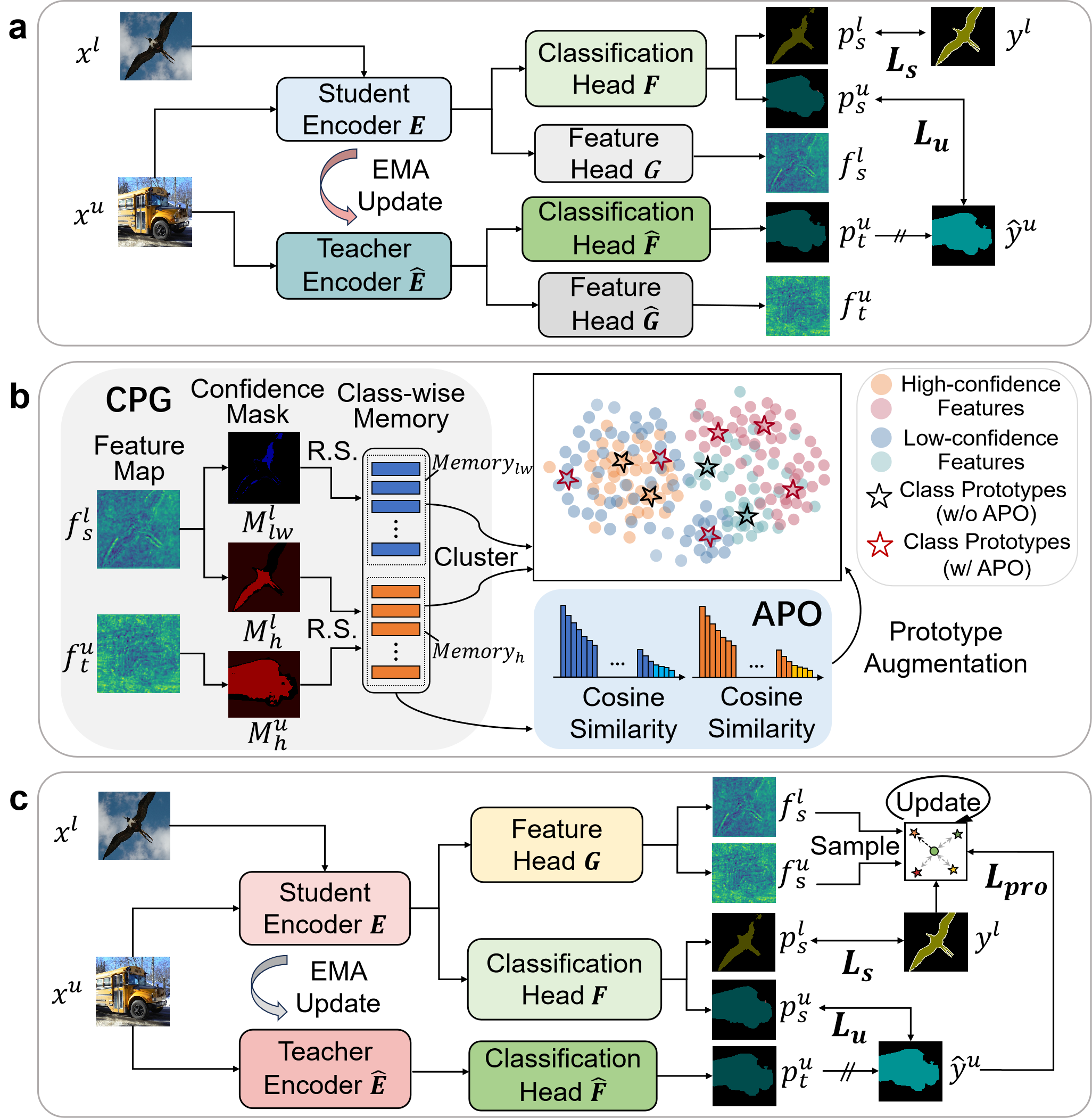}
\caption{An overview of our training framework. (a) presents the pretraining stage of the mean teacher model. (b) shows the proposed boundary-refined prototype generation (BRPG) method with confidence-based prototype generation (CPG) (gray region) and adaptive prototype optimization (APO) (blue region). ``R.S." denotes the random sampling. (c) illustrates the overall training process with the generated prototypes.}

\label{fig3}
\end{figure*}

Fig.~\ref{fig3} illustrates the overall framework of our approach, containing the specific details of these three stages, which will be elaborated in Section \ref{sec:pretraining}, Section \ref{sec:BRPG}, and Section \ref{sec:Training with Prototype-Based Contrastive Learning}, respectively.

\subsection{Pretraining}\label{sec:pretraining}
As shown in Fig. ~\ref{fig3} (a), the pretraining process of the model is conducted based on the mean teacher architecture, in which the student network is optimized through backpropagation, and the weights of the teacher network are updated as the exponential moving average (EMA) \cite{meanteacher} of the student network’s weights. Each network comprises an encoder $E$, a decoder with a classification head $F$, and a feature head $G$.

Given a batch of $B_l$ labeled samples $\left\{\left(x_{i}^{l}, y_{i}^{l}\right)\right\}_{i=1}^{B_{l}}$, the student network is supervised by the following loss function:
\begin{equation}\label{eq:Ls}
L_{s}=\frac{1}{B_{l}} \frac{1}{H W} \sum_{i=1}^{B_{l}} \sum_{j=1}^{H W} l_{c e}\left(p_{s, i j}^{l}, y_{i j}^{l}\right),
\end{equation}
where $p_{s, i j}^{l}=\operatorname{Softmax}\left(F \circ E\left(x_{i j}^{l}\right)\right)$ denotes the predicted probabilities by the student model for the $j$-th pixel of the $i$-th labeled image. Here, $F \circ E$ is the joint mapping of encoder $E$ and classification head $F$. The ground-truth label for this pixel is denoted as $y_{ij}^l$, and the cross-entropy loss of this pixel is computed using the function $l_{ce} (p_{s, i j}^{l}, y_{i j}^{l})$. $H$ and $W$ represent the height and width of the image, respectively.

For unlabeled images $\left\{x_{i}^{u}\right\}_{i=1}^{B_{u}}$, the pseudo-labeling strategy is adopted to extend the supervision of the student model training. In particular, the predictions of the teacher model are used to generate pseudo-labels as follows:
\begin{equation}\label{eq:hat y iju}
\hat{y}_{i j}^{u}=\underset{c}{\operatorname{argmax}}\left(p_{t, i j}^{u}\right), 
\end{equation}
where $p_{t, i j}^{u}=\operatorname{Softmax}\left(\hat{F} \circ \hat{E}\left(x_{i j}^{u}\right)\right)$ represents the softmax probabilities generated by the teacher network. 

To ensure the high quality of pseudo-labels, it is common to introduce a confidence threshold $\eta$ or an entropy threshold $\beta$ to filter out potential noise. For unlabeled images, the pseudo-label mask is defined as follows:
\begin{equation}\label{eq:Miju confidence}
M_{i j}^{u}=\mathbbm{1} \left(\max \left(p_{t, i j}^{u}\right) \geq \eta\right)
\end{equation}
or
\begin{equation}\label{eq:Miju2 entropy}
M_{i j}^{u}=\mathbbm{1} \left(\mathcal{H}\left(p_{t, i j}^{u}\right)<\beta\right),
\end{equation}
where $\mathcal{H}(\cdot)$ represents the pixel-level entropy function. Pixels that satisfy the confidence or entropy condition are assigned a value of 1, while the remaining ones are set as 0. In this way, the unsupervised loss function guided by pseudo-labels can be formulated as:
\begin{equation}\label{eq:Lu}
L_{u}=\frac{1}{M} \sum_{i=1}^{B_{u}} \sum_{j=1}^{H W} l_{c e}\left(p_{s, i j}^{u}, \hat{y}_{i j}^{u}\right) \cdot M_{i j}^{u},
\end{equation}
where $M=\sum_{i=1}^{B_{u}} \sum_{j=1}^{H W} M_{i j}^{u}$ represents the total number of pixels involved in the loss calculation and $p_{s, i j}^{u}=\operatorname{Softmax}\left(F \circ E\left(x_{i j}^{u}\right)\right)$ denotes the student model's predictions for unlabeled data.

Thus, in the pretraining stage, the total loss function can be formulated as follows:  
\begin{equation}\label{eq:L=Ls+Lu}
L=L_{s}+\lambda_{u} L_{u},
\end{equation}
where $\lambda_{u}$ is the weight of unsupervised loss.

Note that during pretraining, the segmentation model also extracts features from both labeled and unlabeled images to facilitate subsequent sampling and prototype generation. Specifically, the features extracted by the student model from the $i$-th labeled image can be represented as $f_{s, i}^{l}=G \circ E\left(x_{i}^{l}\right)$. Meanwhile, to ensure robustness, the features for unlabeled images are consistently obtained from the teacher model, and the resulting feature map is denoted as $f_{t, i}^{u}=\hat{G} \circ \hat{E}\left(x_{i}^{u}\right)$.

\subsection{Boundary-Refined Prototype Generation (BRPG)}\label{sec:BRPG}
In this section, a novel online boundary-refined prototype generation (BRPG) method is proposed to clarify the initial class boundaries of the model. As shown in Fig. ~\ref{fig3} (b), the BRPG consists of two main components, namely, CPG and APO, which will be discussed in Section \ref{sec:CPG} and Section \ref{sec:APO}, respectively.
\subsubsection{Confidence-Based Prototype Generation (CPG)}\label{sec:CPG}
Prototype-based learning \cite{prolearn1,prolearn2} has achieved notable success in semantic segmentation tasks \cite{suc1,suc2,suc3}. The core is to abstract each class as a set of non-learnable prototypes and perform classification by comparing samples to these prototypes. Thus, the initialization strategy for prototypes plays an important role in the classification task.

The experimental results in \cite{badgan} demonstrated a significant improvement in model classification by refining the class boundaries. Inspired by this, we aim to initialize a specific number of prototypes in sparse regions of feature distributions for each class. This method enables the prototypes to capture more challenging samples and refine the class boundaries in prototype-based learning. Fig.~\ref{fig2} illustrates that low-confidence features tend to depart more from the category center, while Fig.~\ref{fig1} shows that low-confidence prototypes are located closer to the class boundaries, which is consistent with our research goal. Consequently, a confidence-based prototype generation (CPG) strategy is proposed, which is depicted in the gray region of Fig.~\ref{fig3} (b).

Specifically, during the sampling stage (with epochs ranging from $T_{S1}$ to $T_{S2}$), for the features $f_s^l$ and $f_t^u$ generated in pretraining, a confidence threshold is introduced to facilitate sampling high- and low-confidence features separately. In addition, to maintain a stable number of samples for each class, two category-wise memory banks (FIFO queues), i.e., ${Memory}_h$ and ${Memory}_{lw}$, are utilized to store these features. Formally, the whole process for high-confidence features can be described as follows:
\begin{equation}\label{eq:Memory h }
    {Memory}_{h}^{(t)}=Q\left({Memory}_{h}^{(t-1)}, S\left(\left\{f_{s, h, i}^{l}\right\}_{i=1}^{B_{l}},\left\{f_{t, h, i}^{u}\right\}_{i=1}^{B_{u}}\right)\right),
\end{equation}
where $f_{s,h,i}^l$ denotes the high-confidence features extracted from the $i$-th labeled image by the student model and $f_{t,h,i}^u$ represents those extracted from the $i$-th unlabeled image by the teacher model. In the case of labeled images, pixels are classified as “high confidence” based on two conditions: (1) the predicted label belongs to the same class as the ground truth, and (2) the softmax probability for the predicted class (i.e., confidence probability) is not less than $\eta_s$. In the experiment, $\eta_s$ is set to 0.8 according to the ablation study. For unlabeled images, high-confidence features are solely needed to have a confidence probability of not less than 0.8 since the ground-truth labels are unavailable.

$S$ denotes the random sampling of high-confidence features with a maximum quantity limit of $(B_l+B_u)*{sample\_num}$, where ${sample\_num}$ is set to 5000. $Q$ represents a first-in-first-out (FIFO) operation performed on storage queues for each class, facilitating the dynamic update of features. 

For low-confidence features, sampling is only performed on labeled images to avoid noise:
\begin{equation}\label{eq:Memory low}
    {Memory}_{l w}^{(t)}=Q\left({Memory}_{l w}^{(t-1)}, \hat{S}\left(\left\{f_{s, l w, i}^{l}\right\}_{i=1}^{B_{l}}\right)\right),
\end{equation}
where $f_{s,lw,i}^l$ represents the low-confidence features of labeled images, which satisfy the condition of correct predictions but have confidence probabilities lower than $\eta_s$. The maximum sampling limit for $ \hat{S}$ is set to ${B_l}*{sample\_num}$. In addition, both memory banks have a storage capacity of $V=30000$ for each class.

After the sampling stage, the K-Means clustering algorithm is applied to both memory banks to generate high- and low-confidence cluster centers for each category, which serve as class prototypes. This can be formulated as follows:
\begin{equation}\label{eq:Rc}
    R_{c}=\left\{\operatorname{Cluster}\left({Memory}_{h, c}, N_{h, c}\right), \operatorname{Cluster}\left({Memory}_{l w, c}, N_{l w, c}\right)\right\}=\left\{r_{c, k}\right\}_{k=1}^{N_{c}},
\end{equation}
where $N_{h,c}$ and $N_{lw,c}$ denote the number of high- and low-confidence cluster centers for class $c$, respectively, so the total number is $N_c=N_{h,c}+N_{lw,c}$.

\subsubsection{Adaptive Prototype Optimization (APO)}\label{sec:APO}
Considering the variation in feature diversity among different categories, assigning an equal number of prototypes to each class may not be the optimal solution. To address this, a simple yet effective technique called adaptive prototype optimization (APO) is proposed, as shown in the blue region of Fig.~\ref{fig3} (b). APO aims to increase the number of cluster centers for classes with more scattered feature distributions, thereby enhancing the refinement of classification boundaries in prototype-based learning. To measure the dispersion of features, two indicators are taken into account: cosine similarity and $l2$ distance.

Specifically, for the cosine similarity indicator, the average cosine similarity between the features and their class center is calculated for each category. The dispersion score for category $c$ is computed as follows:
\begin{equation}\label{eq:D sim alpha c}
    D_{sim, \alpha, c}=\frac{1}{V_{\alpha, c}} \sum_{f \in M e m o r y_{\alpha, c}}\langle f, \bar{f}\rangle,
\end{equation}
where $\langle\cdot, \cdot\rangle$ denotes the similarity metric function and $\alpha\in\left\{h, lw\right\}$ indicates that the dispersion scores of high- and low-confidence features are calculated separately. $V_{\alpha, c}$ is the actual number of stored samples, while $\bar{f}=\frac{1}{V_{\alpha, c}} \sum_{f \in M e m o r y_{\alpha, c}} f $ represents the class center. A lower dispersion score indicates a more scattered feature distribution.

The dispersion scores for all categories can be represented as $D_{\operatorname{sim}, \alpha}=\left\{D_{\operatorname{sim}, \alpha, c}\right\}_{c=1}^{C}$, and an additional cluster center will be added to the categories whose scores are on the bottom $\alpha_t$. Thus, the prototype number $N_{\alpha, c}$ $(\alpha\in\left\{h, lw\right\})$ in Eq.~(\ref{eq:Rc}) can be defined as:
\begin{equation}\label{eq:N erfa c sim}
    N_{\alpha, c}=n_{0 \alpha}+\mathbbm{1}\left(D_{\operatorname{sim}, \alpha, c}<\gamma_{t}\right),
\end{equation}
where $\gamma_t$ denotes the quantile threshold corresponding to $\alpha_t$ ,i.e., $\gamma_{t}=$ $np.\allowbreak percentile$ $\left(D_{\text {sim}, \alpha}, 100 * \alpha_{t}\right)$ with $\alpha_t$ set to 5\% in our implementation. In addition, $n_{0 \alpha}$ is the predefined number of prototypes, which is set to 2 in our experiments.

Similarly, for the $l2$ distance indicator, the dispersion score can be formulated as:
\begin{equation}\label{eq:D l2 alpha c}
    D_{l 2, \alpha, c}=\frac{1}{V_{\alpha, c}} \sum_{f \in M e m o r y_{\alpha, c}}\|f-\bar{f}\|_{2},
\end{equation}
where $\|\cdot\|_{2}$ represents the $l2$ norm. Unlike the case of the cosine similarity indicator, a higher dispersion score based on the $l2$ distance indicates a more scattered feature distribution. Hence, for the dispersion scores $D_{l 2, \alpha}=\left\{D_{l 2, \alpha, c}\right\}_{c=1}^{C}$, only the top $\alpha_t$ categories are eligible to have an additional prototype:
\begin{equation}\label{eq:N alpha c l2}
     N_{\alpha, c}=n_{0 \alpha}+\mathbbm{1}\left(D_{l 2, \alpha, c}>\gamma_{t}\right),
\end{equation}
where the threshold is defined as $\gamma_{t}=np.percentile\left(D_{l 2, \alpha}, 100 *\left(1-\alpha_{t}\right)\right)$.

Note that the cosine similarity indicator exhibits better performance in our method, and therefore, it is adopted by default (refer to Section \ref{sec:ablation study} for the comparison).

\subsection{Training with Prototype-Based Contrastive Learning}\label{sec:Training with Prototype-Based Contrastive Learning}
With the class prototypes generated by BRPG, the semi-supervised model can be regularized by incorporating prototype-based contrastive learning, as illustrated in Fig.~\ref{fig3} (c). The algorithm is introduced as follows:

Due to the dense predictions in semantic segmentation, it is computationally expensive to compare the features of each pixel with prototypes. To address this issue, grid sampling is adopted for labeled images to compute the cosine similarities between the sampled features and prototypes. Particularly, the similarity between the sampled feature $i$ extracted by the feature head and category $c$ is defined as the maximum similarity among all prototypes of category $c$, which can be formulated as:
\begin{equation}\label{eq:s i,c}
     s_{i, c}=\max \left\{<i, r_{c, k}>\right\}_{k=1}^{N_{c}}.
\end{equation}

In this case, the probability of assigning sample $i$ to class $c$ can be estimated in the student model's feature head via:
\begin{equation}\label{eq:p pro i,c}
     p_{i, c}^{p r o}=\frac{\exp \left(s_{i, c} / \tau\right)}{\sum_{t=1}^{C} \exp \left(s_{i, t} / \tau\right)},
\end{equation}
where $C$ represents the number of classes and $\tau$ is the temperature coefficient. Following \cite{pcr}, we set $\tau$ to 0.1.

Therefore, the loss function on labeled data is formulated as follows:
\begin{equation}\label{eq:Lpro l}
\begin{split}
     L_{\text {pro }}^{l}=&-\frac{1}{B_{l} \times M_{l}} \sum_{i=1}^{B_{l}} \sum_{j=1}^{M_{l}} \log p_{i j, y_{i j}^{l}}^{p r o}\\=&-\frac{1}{B_{l} \times M_{l}} \sum_{i=1}^{B_{l}} \sum_{j=1}^{M_{l}} \log \frac{\exp \left(s_{i j, y_{i j}^{l}} / \tau\right)}{\sum_{t=1}^{C} \exp \left(s_{i j, t} / \tau\right)},
\end{split}
\end{equation}
where $M_l$ denotes the number of sampled features per labeled image. Note that the granularity of grid sampling is set to 32, resulting in a total of $32\times32$ samples in each image.

For unlabeled data, to promote the consistency between the two heads, we utilize the pseudo-labels generated by the teacher model’s classification head to guide the prototype-based learning:
\begin{equation}\label{eq:Lpro u}
\begin{split}
     L_{p r o}^{u}=&-\frac{1}{B_{u} \times M_{u}} \sum_{i=1}^{B_{u}} \sum_{j=1}^{M_{u}} \log p_{i j, \hat{y}_{i j}^{u}}^{p r o}\\=&-\frac{1}{B_{u} \times M_{u}} \sum_{i=1}^{B_{u}} \sum_{j=1}^{M_{u}} \log \frac{\exp \left(s_{i j, \hat{y}_{i j}^{u}} / \tau\right)}{\sum_{t=1}^{C} \exp \left(s_{i j, t} / \tau\right)},
\end{split}
\end{equation}
$$\text { s.t. } \max \left(p_{t, i j}^{u}\right) \geq \eta_{t},$$
where $M_u$ denotes the number of randomly sampled features per unlabeled image, with a fixed value of 1000 to match the sampling size of labeled data. Moreover, to mitigate overfitting to the noise in pseudo-labels, the sampling is restricted to pixels whose confidence probabilities are not less than the threshold $\eta_t$.

However, during the training procedure, the confidence of pseudo-labels tends to increase gradually. Based on this intuition, a linear strategy is leveraged to incrementally adjust the threshold $\eta_t$ at each iteration:
\begin{equation}\label{eq:eta t}
     \eta_{t}=\eta_{0}+\left(\eta_{e}-\eta_{0}\right) \frac{{ curr\_iter }}{ { total\_iter }},
\end{equation}
where $\eta_0$ and $\eta_e$ denote the initial and final thresholds, which are set to 0.8 and 0.95. $curr\_iter$ and $total\_iter$ represent the number of completed iterations and the total number of iterations, respectively, after the sampling stage.

Therefore, the overall loss for prototype-based contrastive learning is defined as: 
\begin{equation}\label{eq:L pro}
     L_{p r o}=L_{p r o}^{l}+L_{p r o}^{u}.
\end{equation}

The total loss function with $L_{pro}$ can be formulated as:
\begin{equation}\label{eq:L total final}
     L=L_{s}+\lambda_{u} L_{u}+\lambda_{ {pro }} L_{ {pro }},
\end{equation}
where the weight parameter $\lambda_{pro}$ is set to 1 in the experiment.

Moreover, in our method, the prototypes are dynamically updated with the $M_l+M_u$ features sampled from both labeled and unlabeled images, ensuring their compatibility with the evolving model. For labeled data, each sampled feature is used to update the most similar prototype within the same category. In the case of unlabeled features, the categories of their pseudo-labels are leveraged as references due to the unavailability of ground truths. The update of each prototype can be formally defined as:
\begin{equation}\label{eq:r c,k}
     r_{c, k}^{(t)}=\alpha \cdot r_{c, k}^{(t-1)}+(1-\alpha) \cdot \bar{f}_{c, k},
\end{equation}
where $\bar{f}_{c, k}$ is the mean value of all features assigned to this prototype and the hyper-parameter $\alpha$ controls the update speed, which is set to 0.99.

Finally, the whole pipeline of our method is outlined in Algorithm~\ref{alg:myalgorithm}, with the BRPG process described from line 8 to line 17.

\clearpage
\begin{breakablealgorithm}
\caption{An overview of our approach}
\label{alg:myalgorithm}
\renewcommand{\algorithmicrequire}{\textbf{Input:}}
\renewcommand{\algorithmicensure}{\textbf{Output:}}
\begin{algorithmic}[1]
\REQUIRE $D_{L}$: labeled dataset, $D_{U}$: unlabeled dataset
\ENSURE the parameters of the teacher network \\
\renewcommand{\algorithmicensure}{\textbf{Initialization:}}
\ENSURE  $T$: total number of epochs, $\left[T_{S1}, T_{S2}\right]$: the epoch range for sampling, $n_{iter}$: the number of iterations per epoch, $B_{l}$: batch size of labeled data, $B_{u}$: batch size of unlabeled data, $\lambda_{u}$: weight of unsupervised loss, $\lambda_{pro}$: weight of prototype-based loss \\
\renewcommand{\algorithmicensure}{\textbf{Process:}}
\ENSURE

\FOR{$e \leftarrow 0$ \textbf{to} $T-1$}
    \FOR{$i \leftarrow 0$ \textbf{to} $n_{iter}-1$}
        \STATE Sample a batch of labeled data $\left\{\left(x_{i}^{l}, y_{i}^{l}\right)\right\}_{i=1}^{B_{l}}$ and unlabeled data $\left\{x_{i}^{u}\right\}_{i=1}^{B_{u}}$ from $D_{L} \cup D_{U}$
        \STATE Calculate the supervised loss $L_{S}$ based on Eq.~(\ref{eq:Ls})
        \STATE Calculate the unsupervised loss $L_{u}$ based on Eq.~(\ref{eq:Lu})
        \STATE Calculate the overall loss $L \leftarrow L_{S} + \lambda_{u} L_{u}$
        \STATE Extract features $\left\{f_{s, i}^{l}\right\}_{i=1}^{B_{l}}$ and $\left\{f_{t, i}^{u}\right\}_{i=1}^{B_{u}}$ from both labeled and unlabeled images
        \IF{$e \geq T_{S1}$ \textbf{and} $e < T_{S2}$}
            \STATE Obtain high-confidence features $\left\{f_{s, h, i}^{l}\right\}_{i=1}^{B_{l}}$ and low-confidence features $\left\{f_{s, lw, i}^{l}\right\}_{i=1}^{B_{l}}$ of labeled data via the confidence mask
            \STATE Obtain high-confidence features $\left\{f_{t, h, i}^{u}\right\}_{i=1}^{B_{u}}$ of unlabeled data via the confidence mask
            \STATE Sample and store the high-confidence features in ${Memory}_h$ based on Eq.~(\ref{eq:Memory h })
            \STATE Sample and store the low-confidence features in ${Memory}_{lw}$ based on Eq.~(\ref{eq:Memory low})
        \ELSE
            \IF{$e = T_{S2}$ \textbf{and} $i = 0$}
                \STATE Calculate the prototype number $N_{\alpha, c}$ for each category based on Eq.~(\ref{eq:D sim alpha c}) and Eq.~(\ref{eq:N erfa c sim})
                \STATE Perform K-Means clustering on memory banks to generate prototypes $R_{c}$ based on Eq.~(\ref{eq:Rc})
            \ENDIF
            \STATE Extract features of labeled images from the feature head of the student network and perform grid sampling on them
            \STATE Calculate the loss function of prototype-based contrastive learning $L_{\text{pro}}^{l}$ on labeled data based on Eq.~(\ref{eq:Lpro l})
            \STATE Extract features of unlabeled images from the feature head of the student network and perform random sampling via the confidence threshold $\eta_{t}$
            \STATE Calculate the loss function of prototype-based contrastive learning $L_{\text{pro}}^{u}$ on unlabeled data based on Eq.~(\ref{eq:Lpro u})
            \STATE Calculate the total prototype-based loss $L_{\text{pro}} \leftarrow L_{\text{pro}}^{l} + L_{\text{pro}}^{u}$
            \STATE Update the prototypes $R_{c}$ for each category based on Eq.~(\ref{eq:r c,k})
            \STATE Update the overall loss $L \leftarrow L + \lambda_{\text{pro}} L_{\text{pro}}$
        \ENDIF
        \STATE Update the student network through back-propagation
        \STATE Update the teacher network via EMA of the student network's parameters
    \ENDFOR
\ENDFOR

\end{algorithmic}
\end{breakablealgorithm}

\begin{table}[!ht]
\scriptsize
\caption{Contents of the datasets employed in our experiments.}
\label{tab:dataset}
\centering
\begin{tabular}{|m{3cm}|m{3.2cm}|m{2.8cm}|m{3.3cm}|}
\hline
\textbf{Dataset} & PASCAL VOC 2012 & Cityscapes & MS COCO \\

\hline									
\textbf{Scene} & Vision objects in natural scenes & Urban street scenes & Vision objects in natural scenes \\
\hline

\textbf{Number of classes} & 21 & 19 & 81 \\
\hline

\textbf{Split} & Train/val: 10,582/1,449& Train/val: 2,975/500& Train/val: 118,287/5,000\\
\hline

\textbf{Resolution} & \makecell[l]{Height: up to 500 \\ Width: up to 500} & \makecell[l]{Height: 1024 \\ Width: 2048} & \makecell[l]{Height: up to 640 \\ Width: up to 640} \\
\hline

\textbf{Annotation} & Train: 1,464/9,118 fine/coarse annotations Val: 1,449 fine annotations & Fine annotations & Fine annotations \\
\hline

\textbf{Superclass} & Vehicle, Animal, Household, Person & Flat, Construction, Nature, Vehicle, Sky, Object, Human & Person \& Accessory, Animal, Vehicle, Outdoor Objects, Sports, Kitchenware, Food, Furniture, Appliance, Electronics, Indoor objects \\
\hline

\textbf{Representative class} & Aeroplane, Chair, Dog, Person & Road, Building, Terrain, Bus, Sky, Pole, Rider & Tie, Elephant, Boat, Bench, Skateboard, Spoon, Apple, Couch, Microwave, Laptop, Book \\
\hline

\textbf{Classes per image} & 2.47 & 11.72 & 3.83 \\
\hline
\textbf{Description} & PASCAL VOC 2012 comprises a wide spectrum of visual objects in natural images. & Cityscapes is a benchmark dataset for visual scene understanding, which comprehensively captures the complexity of real-world urban scenes. & MS COCO contains more complex everyday scenes and non-iconic images compared to PASCAL VOC 2012. \\
\hline

\end{tabular}
\end{table}
\normalsize

\section{Experiments}
\subsection{Setup}
In this section, we present the experimental setup of the proposed approach, including six components: datasets, partition protocols, network architectures, training strategies, training details, and evaluation protocols.

\subsubsection{Datasets}
\begin{figure*}[!t]
\centering
\includegraphics[width=0.63\paperwidth]{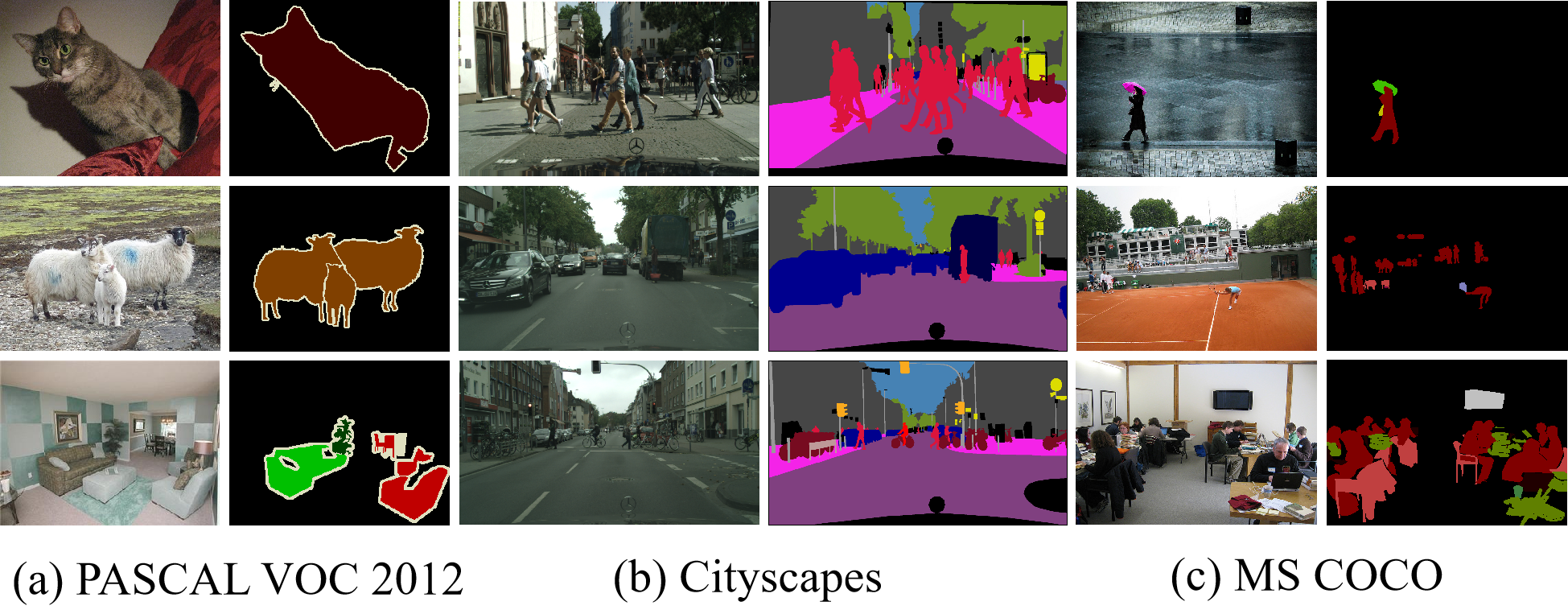}
\caption{Example of (a) PASCAL VOC 2012 \cite{pascal}, (b) Cityscapes \cite{cityscapes} and (c) MS COCO \cite{coco}. The PASCAL dataset mainly comprises iconic-object images (Row 1 and Row 2, objects centered in the image) and iconic-scene images (Row 3, shot from canonical viewpoints and commonly lacking people), while MS COCO features a higher proportion of non-iconic images.
}
\label{fig:dataset}
\end{figure*}

We employ PASCAL VOC 2012 \cite{pascal}, Cityscapes \cite{cityscapes} and MS COCO \cite{coco} to evaluate our approach. Details of these three datasets are presented in Table~\ref{tab:dataset} and Fig.~\ref{fig:dataset}.

\textbf{PASCAL VOC 2012} \cite{pascal} is a benchmark dataset used for visual object segmentation, which contains 20 foreground categories along with a background category. The training set and the validation set consist of 1,464 and 1,449 finely annotated images, respectively. Later, the extra 9,118 training images from the SBD dataset \cite{sbd} are introduced to extend the original dataset, resulting in a total of 10,582 training images. However, the labels obtained from SBD are coarsely annotated and may contain some noise. As a common practice \cite{u2pl,pcr,unimatch}, two settings are utilized for this dataset, i.e., the classic setting and the blended setting. Specifically, the former selects specific proportions of labeled images from the finely annotated set with 1,464 samples, resulting in a low-supervision scenario, while the latter selects from the extended set with all 10,582 labeled images, aiming to demonstrate the model's performance in a high-data regime and its robustness against noise. To evaluate the effectiveness of the proposed method under different annotation qualities and supervision levels, both settings are employed in our experiments.

\textbf{Cityscapes} \cite{cityscapes} is a standard dataset designed for urban scene segmentation. It comprises a total of 2,975 training images and 500 validation images, covering 19 different semantic categories. All images in the dataset have a consistent resolution of 2048 × 1024 pixels.

\textbf{MS COCO} \cite{coco} stands as a challenging benchmark for segmentation, featuring dense annotations across 81 object categories. It exhibits significant scale and diversity, comprising 118,287 training images and 5,000 validation images.

\subsubsection{Partition Protocols}
We employ various partition protocols (i.e., ratios of labeled images) to validate the robustness of the proposed method under varying supervision levels. Specifically, for both settings of PASCAL VOC 2012 and Cityscapes, the proposed approach is evaluated under 1/16, 1/8, 1/4, and 1/2 partition protocols and compared with the state-of-the-art (SOTA) methods under the same data splits from $\text{U}^2\text{PL}$ \cite{u2pl} for a fair comparison. For MS COCO, considering its larger scale, smaller ratios of labeled data are used to evaluate the performance of each method. We utilize the 1/128, 1/64, 1/32, and 1/16 splits from $\text{PC}^{2}\text{Seg}$ \cite{contrast2}.

\subsubsection{Network Architectures}
We employ DeepLabV3+ \cite{deeplabv3+} and SegFormer \cite{segformer} as the basic segmentation architectures. For DeepLabV3+, following the previous works \cite{u2pl,contrast2,pcr}, ResNet-101 \cite{resnet} (for VOC and City) and Xception-65 \cite{xception} (for COCO) pretrained on ImageNet \cite{imagenet} are adopted as the feature backbones. Similar to the inherent classification head, an extra feature head is appended following the ASPP module \cite{Deeplab}. Both heads are composed of two Conv-BN-ReLU-Dropout blocks. However, the feature head differs in mapping the shared features to a 256-dimensional representation space instead of the label space.

For SegFormer, we perform experiments using a Mix Transformer B4 (MiT-B4) \cite{segformer} pretrained on ImageNet \cite{imagenet} as the backbone. Furthermore, a similar Linear-Dropout-Linear structure to the classification head is integrated into the decoder, serving as the feature head with a 256-dimensional mapping.

\subsubsection{Training Strategies}
For PASCAL VOC 2012, we utilize the stochastic gradient descent (SGD) optimizer with the momentum of 0.9 and the weight decay of 0.0001. The initial learning rates of the backbone and decoder are set to 0.001 and 0.01, respectively, and the batch size is $B_l=B_u=16$. The model is trained for 80 epochs with a polynomial learning rate policy: $lr=lr_{init} \cdot\left(1-\frac{i\_{iter}}{{total\_i\_iter}}\right)^{0.9}$. In this dataset, both DeepLabV3+ and SegFormer use the same settings.

For Cityscapes, different configurations are employed for DeepLabV3+ and SegFormer. The former uses an SGD optimizer with an initial learning rate of 0.005, momentum of 0.9, and weight decay of 0.0005. In contrast, SegFormer adopts the AdamW optimizer with betas parameters of (0.9, 0.999) and weight decay of 0.01. The learning rates of the backbone and decoder are set to initial values of 0.00006 and 0.0006, respectively. Both models have a batch size of $B_l=B_u=8$ and are trained for 200 epochs with the identical learning rate policy as applied in the PASCAL dataset.

For the COCO dataset, we conduct experiments with DeepLabV3+, following the same settings as PASCAL VOC 2012, except for the number of training epochs, which is set to 30.

\subsubsection{Training Details}
The setting of pretraining stage in our method follows $\text{U}^2\text{PL}$ \cite{u2pl}. To enhance robustness, data augmentations are applied to both labeled and unlabeled data, including random resizing, random horizontal flipping and random cropping. Specifically, the input images are first randomly resized between 0.5 and 2.0, horizontally flipped with a 0.5 probability, and finally cropped to the designated size. For PASCAL VOC 2012 and MS COCO, the crop size is set to $513\times513$, while for Cityscapes, it is set to $769\times769$. Furthermore, following \cite{unimatch}, strong augmentations, including Color Jitter and CutMix, are applied for the consistency training of unlabeled data. In particular, to ensure high-quality features, feature sampling is conducted after a few epochs of network warm-up. For the PASCAL and COCO dataset, the sampling range is set as $T_{S1}=1$ and $T_{S2}=2$. For Cityscapes, $T_{S1}$ and $T_{S2}$ are set to 3 and 6, respectively, to account for fewer iterations within an epoch. 

Note that all of our experiments are conducted on four NVIDIA Tesla V100 GPUs. However, due to hardware memory limitations, the experimental results can only be reported using mixed precision training \cite{micikevicius2017mixed}.

\subsubsection{Evaluation Protocols}
All evaluations are conducted with the classification head of the teacher network. For PASCAL VOC 2012 and MS COCO, single-scale evaluation is performed on the center-cropped images with crop sizes of $513\times513$ and $641\times641$, respectively, while for Cityscapes, sliding window evaluation is employed with a crop size of $769\times769$ to preserve the original image resolution. Then, the mean of the intersection of union (mIoU) is adopted as the metric to assess the segmentation performance of our model. The reported results are measured on the validation sets of the three datasets.

\subsection{Comparison with State-of-the-Art Methods}

In this section, the proposed method is compared with the SOTA semi-supervised semantic segmentation methods. Additionally, the supervised baselines (denoted as “SupOnly”) trained only on labeled data are also included in the comparison to show the efficacy of our unsupervised algorithms with the extra unlabeled data. To ensure fair comparisons, unless otherwise specified, all the methods are equipped with the DeepLabV3+ network, employing ResNet-101 as the backbone for VOC and Cityscapes, and Xception-65 for COCO. The experimental results on the classic and blended settings of PASCAL VOC 2012 and Cityscapes are presented in Table~\ref{tab1:classic pascal}, Table~\ref{tab2:blended pascal}, and Table~\ref{tab3:cityscapes}, respectively, while the corresponding results for MS COCO are shown in Table \ref{tab:coco result}. Note that all the reported results are the average of three experiments.

\normalsize
\begin{longtable}{>{\raggedright}p{3cm}>{\centering}p{2.4cm}|*{5}{>{\centering\arraybackslash}p{1.7cm}}}
    \caption{ Comparison with the SOTA methods on \textbf{PASCAL VOC 2012} $val$ set under different partition protocols. The methods are trained on the \textbf{\textit{classic}} setting, where labeled images are selected from the high-quality training set with 1,464 samples. The fractions represent the proportion of labeled data utilized for training, followed by the corresponding number of images. ``\dag" indicates that the ``Full" split result is produced according to official code. ``SupOnly" denotes the student model of our method trained solely on labeled images of each partition. ``*" indicates the employment of SegFormer. The best results are in \textbf{bold}, and the second best results are \underline{underlined}.} \label{tab1:classic pascal}\\
    \toprule
    \multirow{2}{*}{\textbf{Method}} & \multirow{2}{*}{\textbf{Venue}} & \textbf{1/16} & \textbf{1/8} & \textbf{1/4} & \textbf{1/2} & \textbf{Full} \\
     & & \textbf{(92)} & \textbf{(183)} & \textbf{(366)} & \textbf{(732)} & \textbf{(1464)}\\
    \midrule
    \endfirsthead

    \toprule
    \multirow{2}{*}{\textbf{Method}} & \multirow{2}{*}{\textbf{Venue}} & \textbf{1/16} & \textbf{1/8} & \textbf{1/4} & \textbf{1/2} & \textbf{Full} \\
     & & \textbf{(92)} & \textbf{(183)} & \textbf{(366)} & \textbf{(732)} & \textbf{(1464)}\\
    \midrule 
    \endhead

    \bottomrule
    \endfoot

    \bottomrule
    \endlastfoot

    MT \cite{meanteacher} & {NeurIPS'17} & 51.72 & 58.93 & 63.86 & 69.51 & 70.96 \\
    CutMix-Seg \cite{semi3} & {BMVC'20} & 52.16 & 63.47 & 69.46 & 73.73 & 76.54 \\
    PseudoSeg \cite{pseudoseg} & {ICLR’21} & 57.60 & 65.50 & 69.14 & 72.41 & 73.23 \\
    $\text{PC}^{2}\text{Seg}$ \cite{contrast2} & {ICCV’21} & 57.00 & 66.28 & 69.78 & 73.05 & 74.15 \\
    $\text{CPS}^{\dag}$ \cite{cps} & {CVPR’21} & 64.07 & 67.42 & 71.71 & 75.88 & 77.47 \\
    CTT \cite{ctt} & {NEUCOM'22} & 64.00 & 71.05 & 72.36 & 76.13 & {-} \\
    ReCo \cite{reco} & {ICLR’22} & 64.78 & 72.02 & 73.14 & 74.69 & {-} \\
    $\text{RC}^{2}\text{L}$ \cite{RC^2L} & {IJCAI'22} & 65.33 & 68.87 & 72.24 & 77.06 & 79.33 \\
    ST++ \cite{st++} & {CVPR’22} & 65.20 & 71.00 & 74.60 & 77.30 & 79.10 \\
    $\text{U}^2\text{PL}$ \cite{u2pl} & {CVPR’22} & 67.98 & 69.15 & 73.66 & 76.16 & 79.49 \\
    PS{-}MT \cite{ps-mt} & {CVPR’22} & 65.80 & 69.58 & 76.57 & 78.42 & 80.01 \\
    PLNS \cite{plns} & {BMVC'22} & 68.60 & 71.40 & 74.40 & 77.50 & 79.50 \\
    PPS \cite{pps} & {ACM MM'22} & 65.96 & 69.32 & 73.45 & 77.63 & {-} \\
    GTA-Seg \cite{gtaseg} & {NeurIPS’22} & 70.02 & 73.16 & 75.57 & 78.37 & 80.47 \\
    PCR \cite{pcr} & {NeurIPS’22} & 70.06 & 74.71 & 77.16 & 78.49 & 80.65 \\
    PRCL \cite{pcrl} & {AAAI'23} & 69.91 & 74.42 & 76.69 & {-} & {-} \\
    $\text{DeS}^{4}$ \cite{des4} & {IJCAI'23} & 68.02 & 72.23 & 74.58 & 77.62 & 80.86 \\
    CCVC \cite{ccvc} & {CVPR’23} & 70.20 & 74.40 & 77.40 & 79.10 & 80.50 \\
    FPL \cite{fpl} & {CVPR’23} & 69.30 & 71.72 & 75.73 & 78.95 & {-} \\
    SemiCVT \cite{semicvt} & {CVPR'23} & 68.56 & 71.26 & 74.99 & 78.54 & 80.32 \\
    UPC \cite{upc} & {ICCV'23} & 71.31 & 73.53 & 76.07 & 77.96 & 80.22 \\
    
    CSS \cite{css} & {ICCV'23} & 68.09 & 71.93 & 74.91 & 77.57 & {-} \\
    Forec \cite{forec} & {arXiv'23} & 71.00 & 74.70 & 77.50 & 78.70 & 81.10 \\
    S4MC \cite{s4mc} & {arXiv'23} & 70.96 & 71.69 & 75.41 & 77.73 & 80.58 \\
    TriKD \cite{TriKD} & {arXiv'23} & 70.29 & 72.53 & 76.80 & 79.71 & 81.13 \\
    MKD \cite{mkd} & {ACM MM'23} & 69.10 & 74.63 & 76.76 & 78.66 & 80.02 \\
    \midrule
    SupOnly & Ours & 45.77 & 54.92 & 65.88 & 71.69 & 72.50 \\
    BRPG & Ours & \underline{73.62}{\color{blue}\footnotesize{$\uparrow${27.85}}} & \underline{76.55}{\color{blue}\footnotesize{$\uparrow${21.63}}} & \underline{78.00}{\color{blue}\footnotesize{$\uparrow${12.12}}} & \underline{79.78}{\color{blue}\footnotesize{$\uparrow${8.09}}} & \underline{81.91}{\color{blue}\footnotesize{$\uparrow${9.41}}} \\
    SupOnly* & Ours & 46.27 & 58.00 & 68.52 & 73.84 & 77.91 \\ 
    BRPG* & Ours & \textbf{75.05}{\color{blue}\footnotesize{$\uparrow${28.78}}} & \textbf{78.35}{\color{blue}\footnotesize{$\uparrow${20.35}}} & \textbf{81.47}{\color{blue}\footnotesize{$\uparrow${12.95}}} & \textbf{83.49}{\color{blue}\footnotesize{$\uparrow${9.65}}} & \textbf{84.57}{\color{blue}\footnotesize{$\uparrow${6.66}}} \\
\end{longtable}
\normalsize

\begin{longtable}{>{\raggedright}p{3cm}>{\centering}p{2.4cm}|*{4}{>{\centering\arraybackslash}p{1.7cm}}}
    \caption{ Comparison with the SOTA methods on \textbf{PASCAL VOC 2012} $val$ set under different partition protocols. The methods are trained on the \textbf{\textit{blended}} setting, where labeled images are selected from the extended training set with 10,582 samples in total.}
    \label{tab2:blended pascal} \\
    \toprule

    \multirow{2}{*}{\textbf{Method}} & \multirow{2}{*}{\textbf{Venue}} & \textbf{1/16} & \textbf{1/8} & \textbf{1/4} & \textbf{1/2} \\
     & & \textbf{(662)} & \textbf{(1323)} & \textbf{(2646)} & \textbf{(5291)} \\

    \midrule
    \endfirsthead

    \toprule
    \multirow{2}{*}{\textbf{Method}} & \multirow{2}{*}{\textbf{Venue}} & \textbf{1/16} & \textbf{1/8} & \textbf{1/4} & \textbf{1/2} \\
     & & \textbf{(662)} & \textbf{(1323)} & \textbf{(2646)} & \textbf{(5291)} \\
    \midrule
    \endhead

    \bottomrule
    \endfoot

    \bottomrule
    \endlastfoot

    MT \cite{meanteacher} & NeurIPS'17 & 70.51 & 71.53 & 73.02 & 76.58 \\
    CutMix-Seg \cite{semi3} & BMVC'20 & 71.66 & 75.51 & 77.33 & 78.21 \\
    CCT \cite{cct} & CVPR'20 & 71.86 & 73.68 & 76.51 & 77.40 \\
    GCT \cite{gct} & ECCV'20 & 70.90 & 73.29 & 76.66 & 77.98 \\
    CPS \cite{cps} & CVPR’21 & 74.48 & 76.44 & 77.68 & 78.64 \\
    AEL \cite{ael} & NeurIPS’21 & 77.20 & 77.57 & 78.06 & 80.29 \\
    $\text{U}^2\text{PL}$ \cite{u2pl} & CVPR’22 & 77.21 & 79.01 & 79.30 & 80.50 \\
    PPS \cite{pps} & ACM MM'22 & 77.29 & 78.22 & 79.51 & 80.46 \\
    GTA-Seg \cite{gtaseg} & NeurIPS’22 & 77.82 & 80.47 & 80.57 & 81.01 \\
    PCR \cite{pcr} & NeurIPS’22 & 78.60 & 80.71 & 80.78 & 80.91 \\
    FST \cite{fst} & NeurIPS’22 & 73.88 & 76.07 & 78.10 & {-} \\
    $\text{DeS}^{4}$ \cite{des4} & IJCAI'23 & 77.28 & 81.02 & 81.61 & \textbf{82.11} \\
    CCVC \cite{ccvc} & CVPR'23 & 76.80 & 79.40 & 79.60 & {-} \\
    AugSeg \cite{augseg} & CVPR'23 & 79.29 & 81.46 & 80.50 & {-} \\
    SemiCVT \cite{semicvt} & CVPR'23 & 78.20 & 79.95 & 80.20 & 80.92 \\
    UPC \cite{upc} & ICCV'23 & 78.53 & 79.92 & 80.36 & 81.05 \\
    CSS \cite{css} & ICCV'23 & 78.73 & 79.54 & 80.82 & 81.06 \\
    Forec \cite{forec} & arXiv'23 & 78.84 & 80.52 & 80.92 & 80.99 \\
    S4MC \cite{s4mc} & arXiv'23 & 78.49 & 79.67 & 79.85 & 81.11 \\
    
    MKD \cite{mkd} & ACM MM'23 & 78.44 & 79.74 & 79.55 & 80.60 \\
    \midrule
    SupOnly & Ours & 67.87 & 71.55 & 75.80 & 77.13 \\
    BRPG & Ours & \underline{79.40}{\color{blue}\footnotesize{$\uparrow${11.53}}} & \underline{81.61}{\color{blue}\footnotesize{$\uparrow${10.06}}} & \underline{81.83}{\color{blue}\footnotesize{$\uparrow${6.03}}} & 80.78{\color{blue}\footnotesize{$\uparrow${3.65}}} \\
    SupOnly* & Ours & 73.47 & 77.70 & 79.32 & 79.56 \\
    BRPG* & Ours & \textbf{83.43}{\color{blue}\footnotesize{$\uparrow${9.96}}} & \textbf{84.53}{\color{blue}\footnotesize{$\uparrow${6.83}}} & \textbf{83.29}{\color{blue}\footnotesize{$\uparrow${3.97}}} & \underline{81.66}{\color{blue}\footnotesize{$\uparrow${2.10}}} \\
\end{longtable}
\normalsize

\begin{longtable}{>{\raggedright}p{3cm}>{\centering}p{2.4cm}|*{4}{>{\centering\arraybackslash}p{1.7cm}}}
    \caption{Comparison with the SOTA methods on \textbf{Cityscapes} $val$ set under different partition protocols. All labeled images are selected from the training set with 2,975 samples in total.}
    \label{tab3:cityscapes} \\
    \toprule
    
    \multirow{2}{*}{\textbf{Method}} & \multirow{2}{*}{\textbf{Venue}} & \textbf{1/16} & \textbf{1/8} & \textbf{1/4} & \textbf{1/2} \\
     & & \textbf{(186)} & \textbf{(372)} & \textbf{(744)} & \textbf{(1488)} \\
    
    \midrule
    \endfirsthead

    \toprule
    \multirow{2}{*}{\textbf{Method}} & \multirow{2}{*}{\textbf{Venue}} & \textbf{1/16} & \textbf{1/8} & \textbf{1/4} & \textbf{1/2} \\
     & & \textbf{(186)} & \textbf{(372)} & \textbf{(744)} & \textbf{(1488)} \\
    \midrule
    \endhead

    \bottomrule
    \endfoot

    \bottomrule
    \endlastfoot
    
    MT \cite{meanteacher} & NeurIPS'17 & 69.03 & 72.06 & 74.20 & 78.15 \\
    CutMix-Seg \cite{semi3} & BMVC'20 & 67.06 & 71.83 & 76.36 & 78.25 \\
    CCT \cite{cct} & CVPR'20 & 69.32 & 74.12 & 75.99 & 78.10 \\
    GCT \cite{gct} & ECCV'20 & 66.75 & 72.66 & 76.11 & 78.34 \\
    CPS \cite{cps} & CVPR’21 & 69.78 & 74.31 & 74.58 & 76.81 \\
    AEL \cite{ael} & NeurIPS’21 & 74.45 & 75.55 & 77.48 & 79.01 \\
    $\text{RC}^{2}\text{L}$ \cite{RC^2L} & IJCAI'22 & {-} & 74.04 & 76.47 & {-} \\
    $\text{U}^2\text{PL}$ \cite{u2pl} & CVPR’22 & 70.30 & 74.37 & 76.47 & 79.05 \\
    PS-MT \cite{ps-mt} & CVPR’22 & {-} & 76.89 & 77.60 & 79.09 \\
    $\text{S4AL}+$ \cite{s4al+} & BMVC'22 & 71.89 & 76.88 & 78.39 & 79.64 \\
    PLNS \cite{plns} & BMVC'22 & 75.70 & 78.00 & 78.70 & {-} \\
    GTA-Seg \cite{gtaseg} & NeurIPS’22 & 69.38 & 72.02 & 76.08 & {-} \\
    PCR \cite{pcr} & NeurIPS’22 & 73.41 & 76.31 & 78.40 & 79.11 \\
    FST \cite{fst} & NeurIPS’22 & 71.03 & 75.36 & 76.61 & {-} \\
    CWC \cite{cwc} & arXiv'22 & 74.50 & 77.00 & 78.60 & {-} \\
    CISC-R \cite{ciscr} & TPAMI'23 & {-} & 75.89 & 77.65 & {-} \\
    $\text{DeS}^{4}$ \cite{des4} & IJCAI'23 & {-} & 75.74 & 77.87 & {-} \\
    FPL \cite{fpl} & CVPR’23 & 75.74 & 78.47 & 79.19 & {-} \\
    UniMatch \cite{unimatch} & CVPR'23 & \underline{76.60} & 77.90 & 79.20 & 79.50 \\
    SemiCVT \cite{semicvt} & CVPR'23 & 72.19 & 75.41 & 77.17 & 79.55 \\
    UPC \cite{upc} & ICCV'23 & 75.31 & 77.35 & 79.03 & 79.62 \\
    3-CPS \cite{3-cps} & ICCV'23 & 75.70 & 77.40 & 78.50 & {-} \\
    
    CSS \cite{css} & ICCV'23 & 74.02 & 76.93 & 77.94 & 79.62 \\
    RRN \cite{rrn} & ICME'23 & 75.42 & 77.34 & 78.38 & 79.09 \\
    Forec \cite{forec} & arXiv'23 & 72.42 & 75.76 & 77.65 & 79.18 \\
    CAFS \cite{cafs} & arXiv'23 & 76.30 & 77.80 & 79.40 & {-} \\
    S4MC \cite{s4mc} & arXiv'23 & 75.03 & 77.02 & 78.78 & 78.86 \\
    TriKD \cite{TriKD} & arXiv'23 & 72.70 & 76.44 & 78.01 & 79.12 \\
    
    DSSN \cite{dssn} & ACM MM'23 & 76.52 & 78.18 & 78.62 & 79.58 \\
    \midrule
    SupOnly & Ours & 65.74 & 72.53 & 74.43 & 77.83 \\
    BRPG & Ours & 76.37{\color{blue}\footnotesize{$\uparrow${10.63}}} & \underline{78.56}{\color{blue}\footnotesize{$\uparrow${6.03}}} & \underline{79.46}{\color{blue}\footnotesize{$\uparrow${5.03}}} & \underline{80.26}{\color{blue}\footnotesize{$\uparrow${2.43}}} \\
    SupOnly* & Ours & 69.43 & 74.03 & 76.93 & 79.18 \\
    BRPG* & Ours & \textbf{76.84}{\color{blue}\footnotesize{$\uparrow${7.41}}} & \textbf{79.65}{\color{blue}\footnotesize{$\uparrow${5.62}}} & \textbf{80.37}{\color{blue}\footnotesize{$\uparrow${3.44}}} & \textbf{80.92}{\color{blue}\footnotesize{$\uparrow${1.74}}} \\
\end{longtable}

\normalsize

\normalsize

\begin{longtable}{>{\raggedright}p{3cm}>{\centering}p{2.4cm}|*{4}{>{\centering\arraybackslash}p{1.7cm}}}
    \caption{Comparison with the SOTA methods on \textbf{MS COCO} $val$ set under different partition protocols. The original $train$ set of 118,287 images is split into labeled subsets ranging from 1/128 to 1/16.}
    \label{tab:coco result} \\
    \toprule

    \multirow{2}{*}{\textbf{Method}} & \multirow{2}{*}{\textbf{Venue}} & \textbf{1/128} & \textbf{1/64} & \textbf{1/32} & \textbf{1/16}  \\
     & & \textbf{(925)} & \textbf{(1849)} & \textbf{(3697)} & \textbf{(7393)} \\
     
    \midrule
    \endfirsthead

    \toprule
    \multirow{2}{*}{\textbf{Method}} & \multirow{2}{*}{\textbf{Venue}} & \textbf{1/128} & \textbf{1/64} & \textbf{1/32} & \textbf{1/16}  \\
     & & \textbf{(925)} & \textbf{(1849)} & \textbf{(3697)} & \textbf{(7393)} \\
    \midrule
    \endhead

    \bottomrule
    \endfoot

    \bottomrule
    \endlastfoot

    PseudoSeg \cite{pseudoseg} & ICLR’21 & 39.11 & 41.75 & 43.64 & {-} \\
    $\text{PC}^{2}\text{Seg}$ \cite{contrast2} & ICCV’21 & 40.12 & 43.67 & 46.05 & \underline{48.07} \\
    MKD \cite{mkd} & ACM MM'23 & \textbf{42.32} & \underline{45.50} & \underline{47.25} & {-} \\
    
    \midrule
    SupOnly & Ours & 33.60 & 37.80 & 42.24 & 45.12 \\
    BRPG & Ours & \underline{41.73}{\color{blue}\footnotesize{$\uparrow${8.13}}} & \textbf{45.91}{\color{blue}\footnotesize{$\uparrow${8.11}}} & \textbf{50.55}{\color{blue}\footnotesize{$\uparrow${8.31}}} & \textbf{53.73}{\color{blue}\footnotesize{$\uparrow${8.61}}}
\end{longtable}

\normalsize

\textbf{Results on PASCAL VOC 2012 Dataset.} Table~\ref{tab1:classic pascal} presents the comparison results on the classic setting of PASCAL VOC 2012. With the utilization of DeepLabV3+, our method exhibits substantial improvements over the supervised baseline by +27.85\%, +21.63\%, +12.12\%, +8.09\%, and +9.41\% under the five label partitions. Moreover, compared with the previous best prototype-based PCR method \cite{pcr}, our approach still yields superior performance, with gains of +3.56\%, +1.84\%, +0.84\%, +1.29\%, and +1.26\% under each partition. This confirms that the proposed end-to-end method can completely achieve and even surpass the existing method with offline prototype generation. In addition, our method outperforms the previous best results across all partitions, particularly on low-data regimes, e.g., +2.31\% and +1.84\% under 1/16 and 1/8 partitions, respectively. Moreover, our approach achieves remarkable improvements with SegFormer, outperforming the baseline model by +28.78\%, +20.35\%, +12.95\%, +9.65\%, and +6.66\% across the five partitions. Surprisingly, the model performance with SegFormer significantly surpasses that with DeepLabV3+, demonstrating the robustness and scalability of our approach.

Table~\ref{tab2:blended pascal} reports the comparison results on the blended setting. With DeepLabV3+, our proposed approach consistently outperforms the supervised baseline by a large margin. For instance, our method surpasses the SupOnly baseline by over 10\% in mIoU under both 1/16 and 1/8 partitions. Compared with the prototype-based method PCR \cite{pcr}, our approach performs better under 1/16, 1/8, and 1/4 partitions, with improvements of +0.8\%, +0.9\%, and +1.05\%, respectively, and achieves a comparable result under 1/2 label partition. In addition, the model with SegFormer significantly outperforms the previous best methods under most partitions.

However, the performance of our method appears to drop under high-data partitions. This may be attributed to the reduced number of unlabeled images, leading to fewer iterations per epoch (as the number of epochs is calculated based on the iterations of unlabeled images). Furthermore, we have observed that the network performance does not reach saturation throughout the whole training process. Hence, a better result may be obtained with a longer training process.

\textbf{Results on Cityscapes Dataset.} Table~\ref{tab3:cityscapes} illustrates the comparison results on Cityscapes. Our approach achieves remarkable performance gains over the SupOnly approach by +10.63\%, +6.03\%, +5.03\%, and +2.43\% under the 1/16, 1/8, 1/4, and 1/2 partition protocols, respectively. In addition, our method outperforms PCR \cite{pcr} under all partitions by +2.96\%, +2.25\%, +1.06\%, and +1.15\%, respectively. Notably, our proposed method achieves superior performance across most partitions. Even when compared to the latest SOTA methods such as UniMatch \cite{unimatch} and DSSN \cite{dssn}, our approach outperforms them under the 1/8, 1/4, and 1/2 partitions, with only a slight disadvantage under the 1/16 partition. However, when employing SegFormer, the proposed method achieves optimal results across all partitions. This demonstrates the effectiveness and robustness of our approach across different network architectures.

\begin{figure*}[!t]
\centering
\includegraphics[width=0.5\paperwidth]{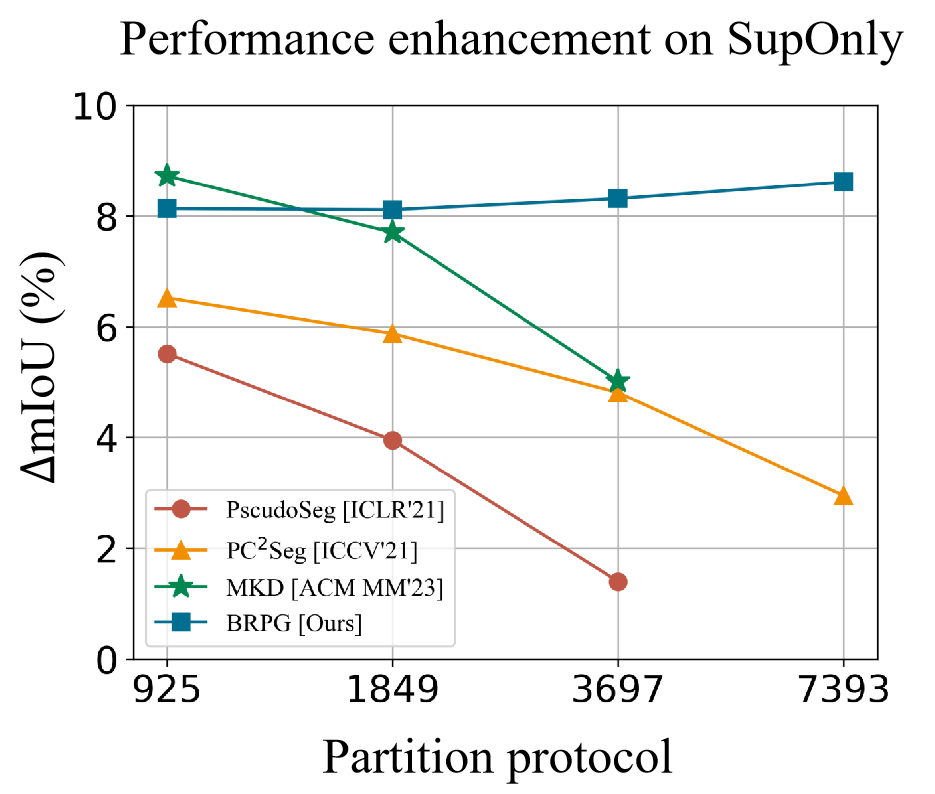}
\caption{The performance improvement of the state-of-the-art methods on \textbf{MS COCO} \textit{val} set.}
\label{fig-coco-perform}
\end{figure*}

\textbf{Results on MS COCO Dataset.} Table~\ref{tab:coco result} illustrates the model performance using DeepLabV3+ on COCO $val$ set. As observed, the proposed BRPG method outperforms all available methods on the three high-data regimes. Notably, our approach consistently demonstrates an improvement of over 8\% in mIoU compared to the SupOnly baseline across all partitions. In contrast, the performance gains achieved by the three compared methods appear to be limited or even reach a saturation point on high-data partitions (as shown in Fig.~\ref{fig-coco-perform}). The experimental results confirm the efficacy and scalability of BRPG on complex large-scale datasets.

\subsection{Ablation Study}\label{sec:ablation study}
To investigate the effectiveness of the proposed modules and the impact of various configurations, ablation studies are conducted on both the PASCAL dataset with 366 and 1323 splits and the Cityscapes dataset with 372 and 744 splits. DeepLabV3+ \cite{deeplabv3+} is adopted as the default segmentation network.

\textbf{Effectiveness of Components.} To manifest the contribution of each component, a series of ablation experiments are conducted, as shown in Table~\ref{tab4:component}. Among them, two baseline models, trained with SupOnly (Experiment I) and pseudo-labeling (Experiment II), are employed for comparison. In Experiment III, the proposed end-to-end prototype-based contrastive learning is introduced to the semi-supervised model with random sampling and prototype generation. This implementation leads to performance improvements of +1.62\% and +1.11\% under 366 and 1323 partitions of the PASCAL dataset, respectively. Similarly, for the Cityscapes dataset, the performance is improved by +1.60\% and +0.81\% under 372 and 744 partitions, respectively. Experiment IV demonstrates the effect of incorporating the confidence-based prototype generation (CPG) strategy, which further improves the performance by +0.62\%, +0.21\%, +0.55\% and +0.74\% under each partition. This can be attributed to the scheme of separate sampling and clustering based on high and low confidences. Finally, in Experiment V, the adaptive prototype optimization (APO) method is leveraged to perform prototype augmentation for categories with dispersed feature distributions, resulting in gains of +0.61\%, +0.41\%, +0.54\% and +0.47\% on the four splits. It is noteworthy that the proposed novel BRPG (CPG+APO) approach exhibits significant improvements over Model III, with total gains of +1.23\%, +0.62\%, +1.09\% and +1.21\% under the four partitions, respectively. This observation indicates the importance of refining initial class boundaries in prototype-based learning.

\begin{table}[!ht]
    \caption{Ablation study on different components of the proposed approach. $L_{sup}$: Supervised training with labeled data. $L_{unsup}$: Unsupervised training with pseudo-labels. $L_{pro}$: Plain prototype-based contrastive learning with random sampling and direct K-Means clustering. CPG: Confidence-based prototype generation. APO: Adaptive prototype optimization.}
    \label{tab4:component}
    \centering
    \begin{tabular}{c|c|c|c|c|c|cc|cc}
    \toprule
        \multirow{2}{*}{~} & \multirow{2}{*}{$L_{sup}$} & \multirow{2}{*}{$L_{unsup}$} & \multirow{2}{*}{$L_{pro}$} & \multirow{2}{*}{CPG} & \multirow{2}{*}{APO} & 
        \multicolumn{2}{c|}{VOC} & \multicolumn{2}{c}{City} \\
        & ~ & ~ & ~ & ~ & ~ & 366 & 1323 & 372 & 744 \\ 
        
        \midrule
        I & \checkmark & ~ & ~ & ~ & ~ & 65.88 & 71.55 & 72.53 & 74.43 \\ 
        II & \checkmark & \checkmark & ~ & ~ & ~ & 75.15 & 79.88 & 75.87 & 77.44 \\ 
        III & \checkmark & \checkmark & \checkmark & ~ & ~ & 76.77 & 80.99 & 77.47 & 78.25\\ 
        IV & \checkmark & \checkmark & \checkmark & \checkmark & ~ & 77.39 & 81.20 & 78.02 & 78.99 \\ 
        V & \checkmark & \checkmark & \checkmark & \checkmark & \checkmark & \textbf{78.00} & \textbf{81.61} & \textbf{78.56} & \textbf{79.46}\\ \bottomrule
    \end{tabular}
\end{table}

\begin{table}[!ht]
\caption{Ablation study on the selection of feature sampling range. The base sampling span for Cityscapes is set to 3 epochs for its fewer iterations per epoch.}
\label{tab:sampling range}
\centering
\begin{tabular}{p{2cm}>{\centering\arraybackslash}p{3.5cm}p{2cm}>{\centering\arraybackslash}p{3cm}}
\toprule
$\left[T_{S1}, T_{S2}\right]$ & VOC (366) & $\left[T_{S1}, T_{S2}\right]$ & City (372) \\
\midrule

$\left[0, 1\right]$ & 77.83 & $\left[0, 3\right]$ & 78.22 \\

$\left[1, 2\right]$ & \textbf{78.00} & $\left[3, 6\right]$ & \textbf{78.56} \\

$\left[2, 3\right]$ & 77.25 & $\left[6, 9\right]$ & 77.66 \\

$\left[5, 6\right]$ & 77.04 & $\left[15, 18\right]$ & 77.81 \\

$\left[0, 2\right]$ & 77.59 & $\left[0, 6\right]$ & 78.05 \\

$\left[1, 3\right]$ & 76.97 & $\left[3, 9\right]$ & 78.35 \\

$\left[1, 6\right]$ & 77.38 & $\left[3, 18\right]$ & 77.93 \\

$\left[1, 11\right]$ & 77.48 & $\left[3, 33\right]$ & 77.54 \\
\bottomrule

\end{tabular}
\end{table}

\textbf{Selection of feature sampling range.} The epoch range for feature sampling impacts the quality of prototype generation. Specifically, insufficiently trained features may lack class semantics, while overtraining may lead to features clustering around the semantic centers but distant from the class boundaries, ultimately compromising the model performance. Therefore, the ablation experiments on various starting points and spans are conducted to determine the optimal sampling range, as illustrated in Table~\ref{tab:sampling range}. It can be observed that setting $\left[T_{S1}, T_{S2}\right]$ to $\left[1, 2\right]$ for PASCAL VOC 2012 and $\left[3, 6\right]$ for Cityscapes yields the superior performance.

\begin{table}[!ht]
\caption{Ablation study on the value of confidence threshold $\eta_s$ during the sampling stage.}
    \label{tab5:confidence threshold}
    \centering
    \begin{tabular}{c|ccccccc}
    \toprule
        $\eta_s$ & 0.7 & 0.75 & 0.8 & 0.85 & 0.9 & 0.95 & $\text{EMA}_{0.7\text{-}0.9}$  \\ \midrule
        VOC (366) & 77.59 & 77.69 & \textbf{78.00} & 77.50 & 77.86 & 77.51 & \textbf{78.14}  \\ \midrule
        City (372) & 78.45 & 78.25 & \textbf{78.56} & 77.72 & 77.89 & 77.51 & \textbf{78.80} \\ \bottomrule
    \end{tabular}
\end{table}

\textbf{Value of the confidence threshold $\eta_s$.} Based on the confidence threshold $\eta_s$, the separate sampling of high- and low-confidence features constitutes a fundamental element of our method. To explore the impact of $\eta_s$ on model performance, ablation studies are conducted with its different values, as shown in Table~\ref{tab5:confidence threshold}. It can be observed that within the range of [0.7, 0.95], $\eta_s=0.8$ yields the best results on both datasets.

Moreover, considering the performance variations across different classes, we also explore the effect of setting class-specific confidence thresholds. Specifically, an initial threshold of 0.8 is assigned to each class, with upper and lower bounds of 0.9 and 0.7, respectively. Subsequently, the thresholds are updated during the sampling stage using the EMA \cite{meanteacher} strategy with a decay rate of 0.999 based on the mean confidence predicted for each category. As presented in Table~\ref{tab5:confidence threshold}, this approach exhibits marginal improvements on both datasets compared to the model with $\eta_s=0.8$. Therefore, future work could focus on identifying optimal solutions for class-specific confidence thresholding.

\begin{figure*}[t]
\centering
\subfigure[PASCAL VOC 2012]{\includegraphics[width=0.3\paperwidth]{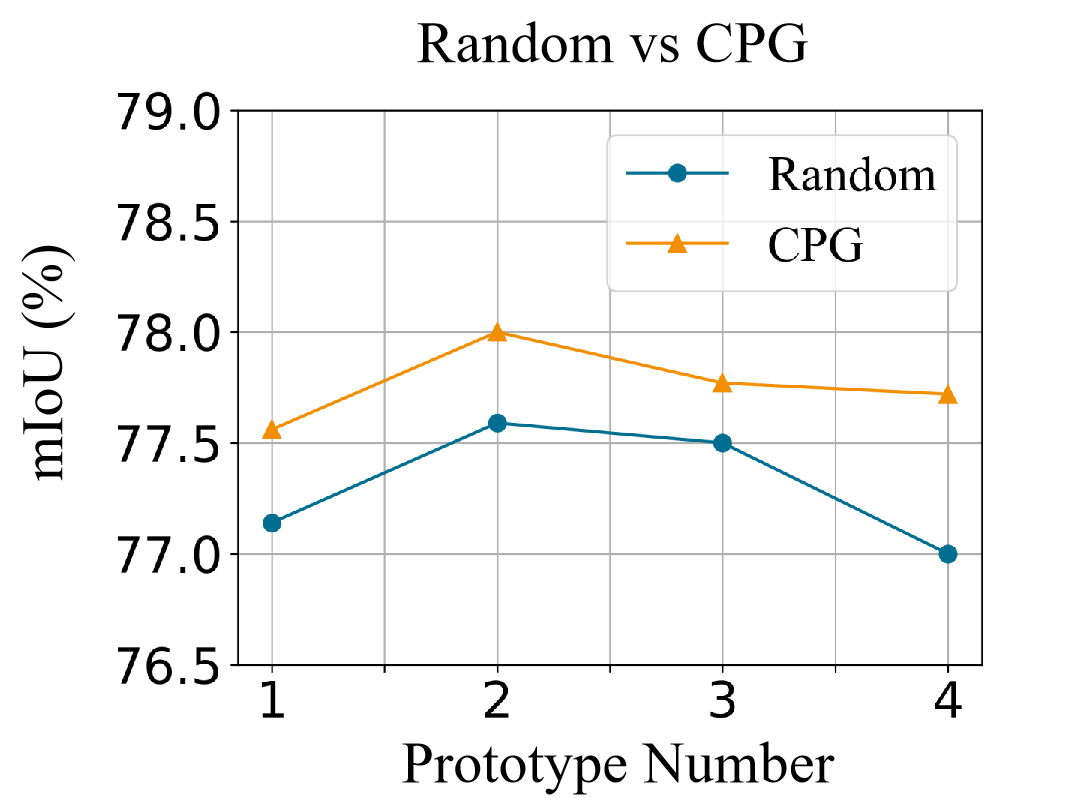}}
\hspace{5mm}
\subfigure[Cityscapes]
{\includegraphics[width=0.3\paperwidth]{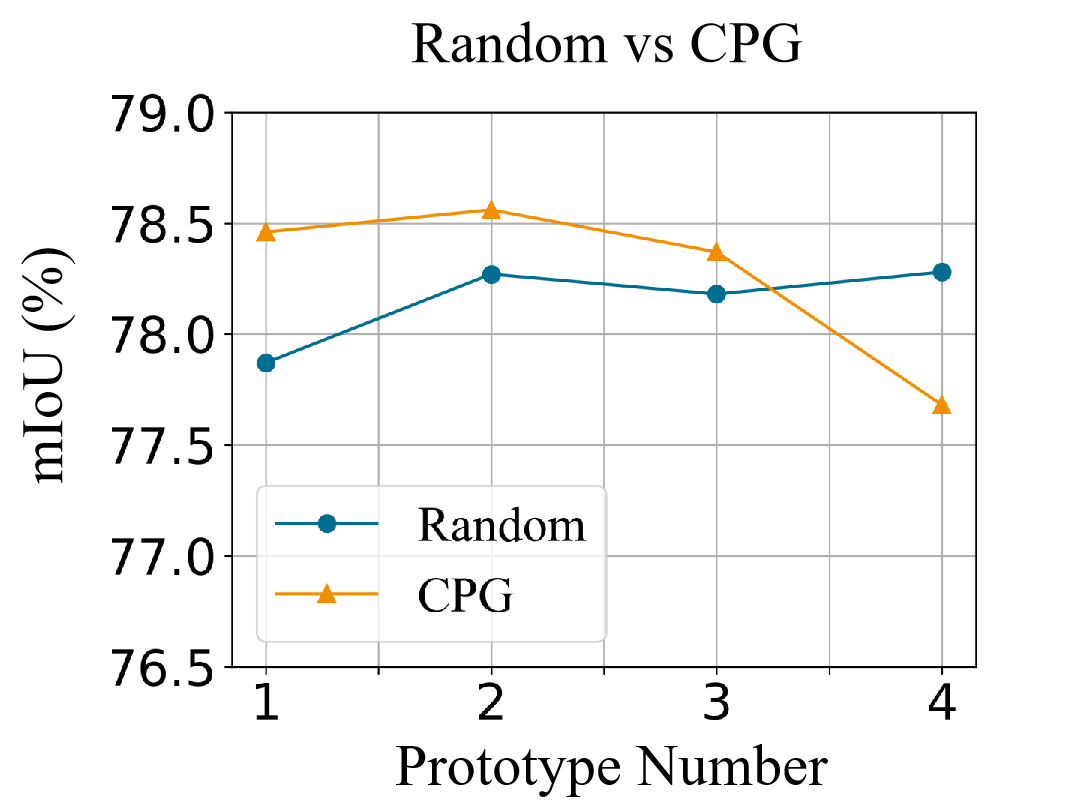}}\\
\caption{Ablation studies on prototype generation strategies with varying numbers. The experiments are conducted on the 366 split of PASCAL VOC 2012 and the 372 split of Cityscapes. ``Random" refers to random sampling and direct K-Means clustering. Note that ``Prototype Number" for ``Random" indicates prototypes per category, while for ``CPG", it represents high- and low-confidence prototypes separately for each category. Despite this difference, CPG outperforms the baseline across most prototype number settings.}
\label{fig4}
\end{figure*}

\begin{figure*}[t]
\centering
\subfigure[PASCAL VOC 2012]{\includegraphics[width=0.3\paperwidth]{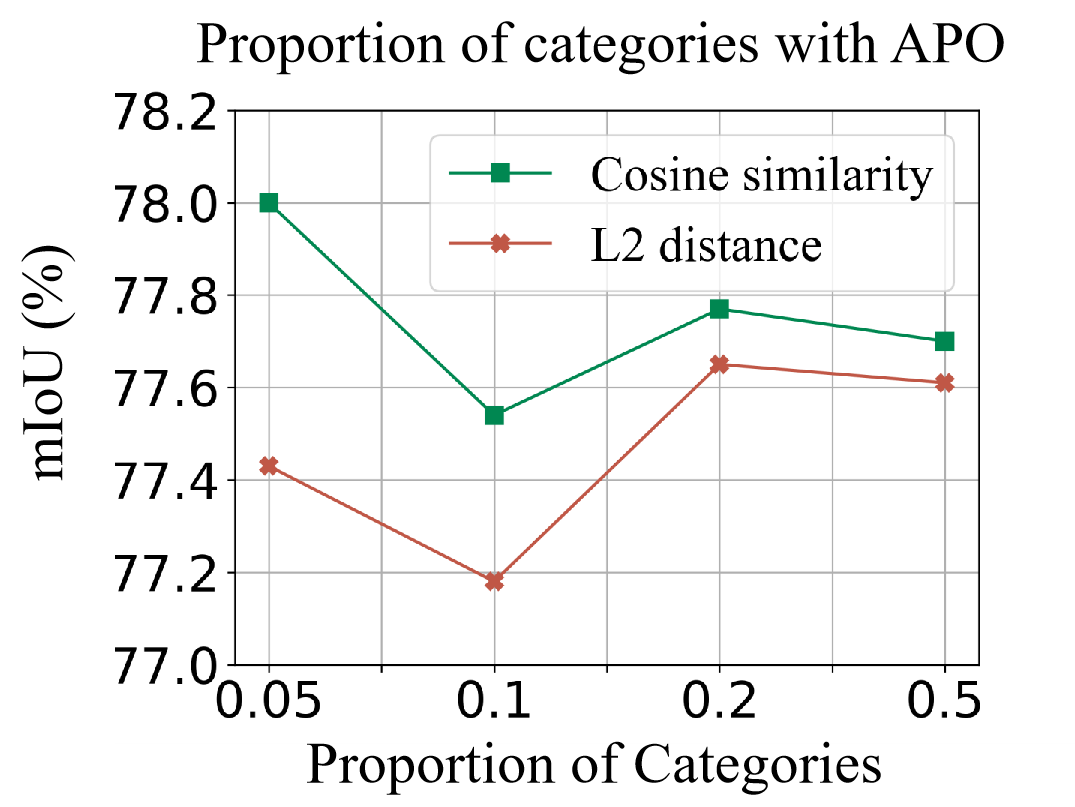}}
\hspace{5mm}
\subfigure[Cityscapes]
{\includegraphics[width=0.3\paperwidth]{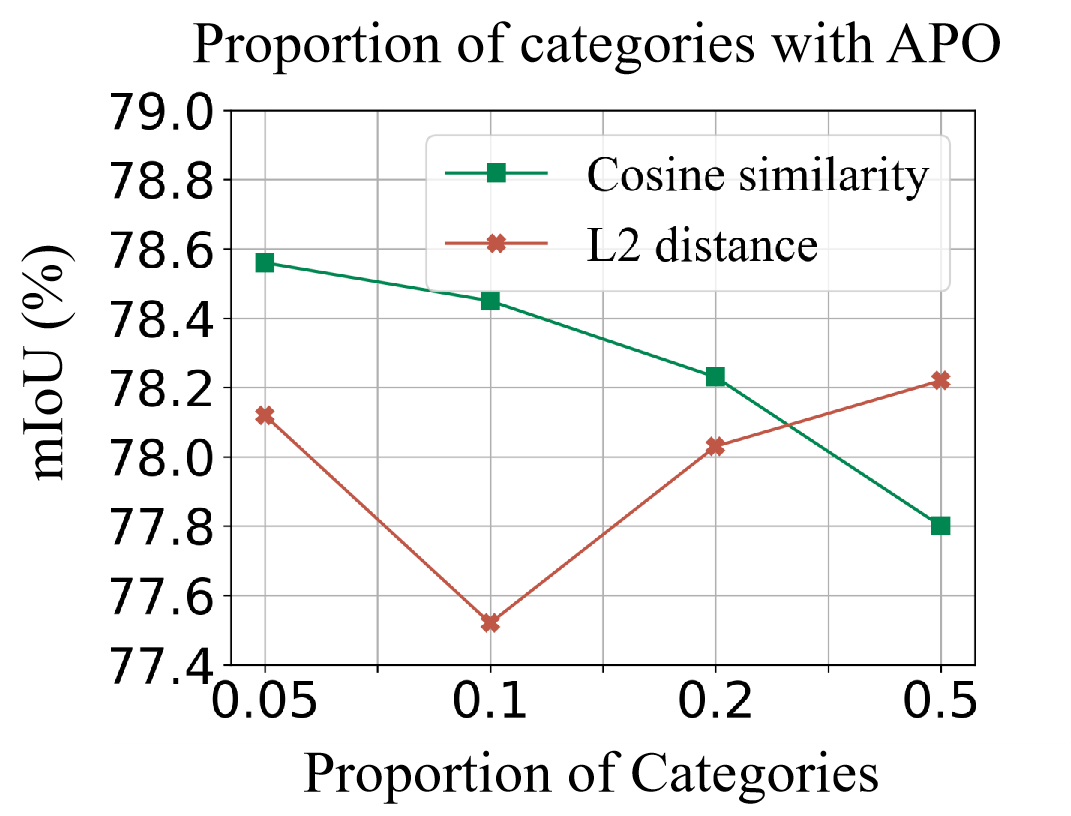}}\\
\caption{Ablation studies on APO under varying class proportions using cosine similarity and $l2$ distance indicators. All the presented results are obtained using the 366 split of PASCAL VOC 2012 and the 372 split of Cityscapes.
}
\label{fig-APO ablation study}
\end{figure*}

\textbf{Strategies for prototype generation.} Fig.~\ref{fig4} presents the impacts of prototype generation strategies and the number of prototypes. For the PASCAL dataset, it can be observed that the proposed confidence-based prototype generation (CPG) method consistently outperforms the baseline method, which involves random feature sampling and direct K-Means clustering. In addition, optimal performance can be obtained when setting the number of high- and low-confidence prototypes to 2 for each category. For Cityscapes, CPG exhibits superiority over the baseline under most settings, achieving satisfactory performance with a prototype number of 1 and 2. To attain optimal performance on both datasets, we set the prototype number of CPG to 2 in our experiments.

\textbf{Proportion of categories with APO.} APO aims to increase the number of cluster centers for categories with scattered feature distributions. Hence, it is crucial to select a reliable indicator that measures the dispersion of features. As mentioned above, two indicators are taken into account, i.e., cosine similarity and $l2$ distance. The ablation studies on the class proportions for implementing APOs are illustrated in Fig.~\ref{fig-APO ablation study}. For both datasets, cosine similarity proves to be a more suitable indicator, as it demonstrates overall superior performance compared to $l2$ distance. The optimal performance can be reached when the class proportion is set to 5\%.

\begin{table}[!ht]
\caption{Ablation study on different configurations for the sampling stage. Baseline (Ours): Feature head is not pretrained. Experiment I: Pretrain the feature head with labeled data before sampling. Experiment II: Extract features from the classification head. Experiment III: Extract features from the ASPP module, discarding the feature head.}
\label{tab6}
    \centering
    \begin{tabular}{c|cccc}
    \toprule
        Split & Baseline (Ours) & Experiment I & Experiment II & Experiment III \\ \midrule
        366 & \textbf{78.00} & 77.24 & 77.60 & 77.17  \\ \midrule
        1323 & 81.61 & \textbf{81.81} & 81.23 & 80.81  \\ \bottomrule
    \end{tabular}
\end{table}

\begin{table}[!ht]
\caption{Scalability validation on the \textbf{\textit{classic}} setting of \textbf{PASCAL VOC 2012}. $\dagger$: Results reported in \cite{unimatch}. $\ddag$: Result of our implementation on the 1464 split using official code.}
\label{tab7}
    \centering
     \begin{tabular}{r|*{5}{>{\centering\arraybackslash}p{1.6cm}}}
    \toprule
        Method & 92 & 183 & 366 & 732 & 1464  \\ \midrule
        $\text{FixMatch}^{\dagger}$ \cite{fixmatch} & 63.90 & 73.00 & 75.50 & 77.80 & 79.20  \\ 
        FixMatch + BRPG & \textbf{67.45}{\color{blue}\footnotesize{$\uparrow${3.55}}}& \textbf{74.71}{\color{blue}\footnotesize{$\uparrow${1.71}}}& \textbf{76.47}{\color{blue}\footnotesize{$\uparrow${0.97}}}& \textbf{78.57}{\color{blue}\footnotesize{$\uparrow${0.77}}}& \textbf{79.70}{\color{blue}\footnotesize{$\uparrow${0.50}}} \\ \midrule
        $\text{CPS}^{\ddag}$ \cite{cps} & 64.07 & 67.42 & 71.71 & 75.88 & 77.47  \\ 
        CPS + BRPG & \textbf{67.85}{\color{blue}\footnotesize{$\uparrow${3.78}}}& \textbf{70.80}{\color{blue}\footnotesize{$\uparrow${3.38}}}& \textbf{74.63}{\color{blue}\footnotesize{$\uparrow${2.92}}}& \textbf{77.85}{\color{blue}\footnotesize{$\uparrow${1.97}}}& \textbf{79.15}{\color{blue}\footnotesize{$\uparrow${1.68}}} \\

        \bottomrule
    \end{tabular}
\end{table}

\textbf{Configurations for the sampling stage.} Note that the feature head of the proposed method is not pretrained before the sampling stage, so the extracted features solely rely on the shared features and the initial parameters of this subnet. To examine the effect of the feature head and sampled features on the final results, we conduct experiments under the following settings: (1) Pretrain the feature head with labeled data before the sampling stage. (2) Extract features from the classification head. (3) Extract 512-dimensional features from the output of the ASPP module, discarding the feature head. Table~\ref{tab6} presents the experimental results on PASCAL VOC 2012. It can be observed that network pretraining (Experiment I) or the use of discriminative features (Experiment II) does not lead to significant performance improvements and may even cause slight degradation. Additionally, extracting features from ASPP for prototype-based training (Experiment III) results in inferior performance, which indicates the necessity of introducing the feature head. Consequently, the baseline setting is adopted as the default in our experiments.

\textbf{More ablation studies.} We provide additional ablation experiments to illustrate the supplementary process of parameter settings. For detailed information, refer to \ref{appendix: A more ablation study}.

\subsection{Scalability Validation}
Moreover, to validate its scalability, the proposed method is applied to two other prevalent semi-supervised frameworks, FixMatch \cite{fixmatch} and CPS \cite{cps}, with DeepLabV3+ as the basic segmentation architecture. Similar to the utilization of the mean teacher \cite{meanteacher}, an additional feature head is incorporated into the segmentation network for the implementation of prototype generation and contrastive learning without altering the inherent training procedure. The combined frameworks are evaluated on the classic setting of PASCAL VOC 2012, as shown in Table~\ref{tab7}. It is clear that both of the combined frameworks achieve consistent performance gains over the baselines, demonstrating the remarkable generalization capability of our approach.

\begin{figure}[htbp]

	\centering
	\subfigure[SupOnly]{
		\begin{minipage}[t]{0.49\linewidth}
		\centering
			\includegraphics[width=1\linewidth]{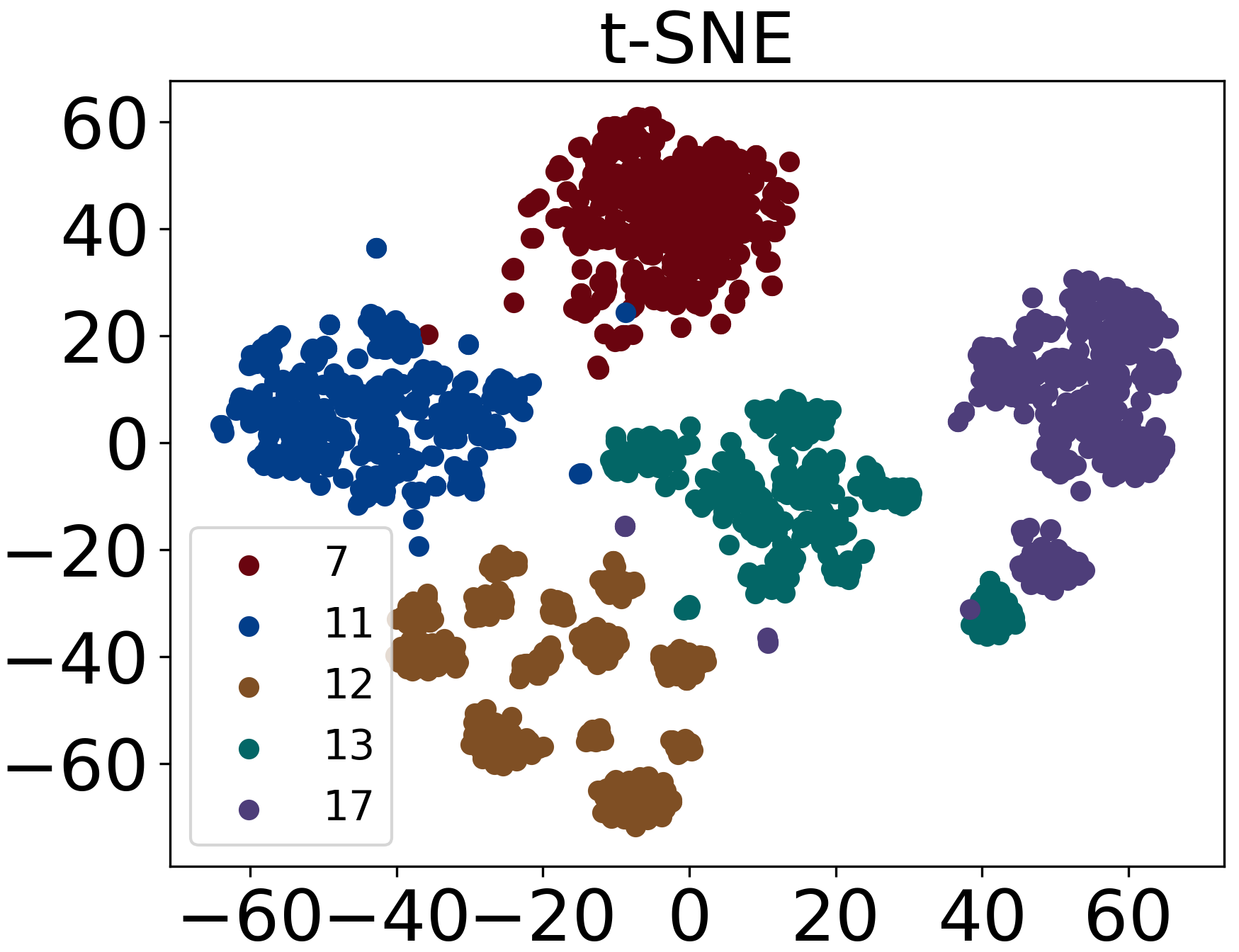}

		\end{minipage}
	}%
	\subfigure[$L_{unsup}$]{
		\begin{minipage}[t]{0.49\linewidth}
		\centering
			\includegraphics[width=1\linewidth]{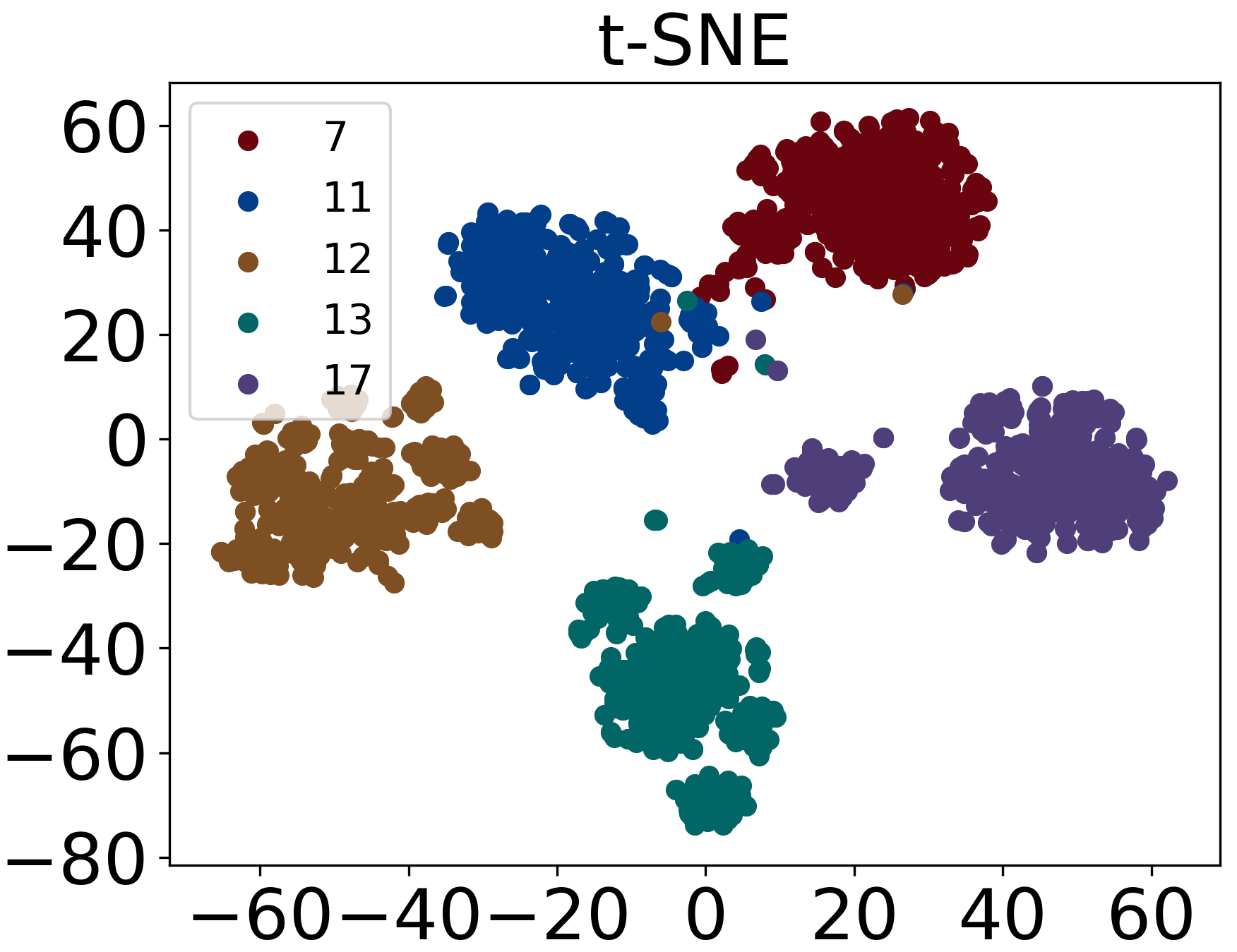}

		\end{minipage}
	}
	
	\subfigure[Plain $L_{pro}$]{
		\begin{minipage}[t]{0.49\linewidth}
		\centering
			\includegraphics[width=1\linewidth]{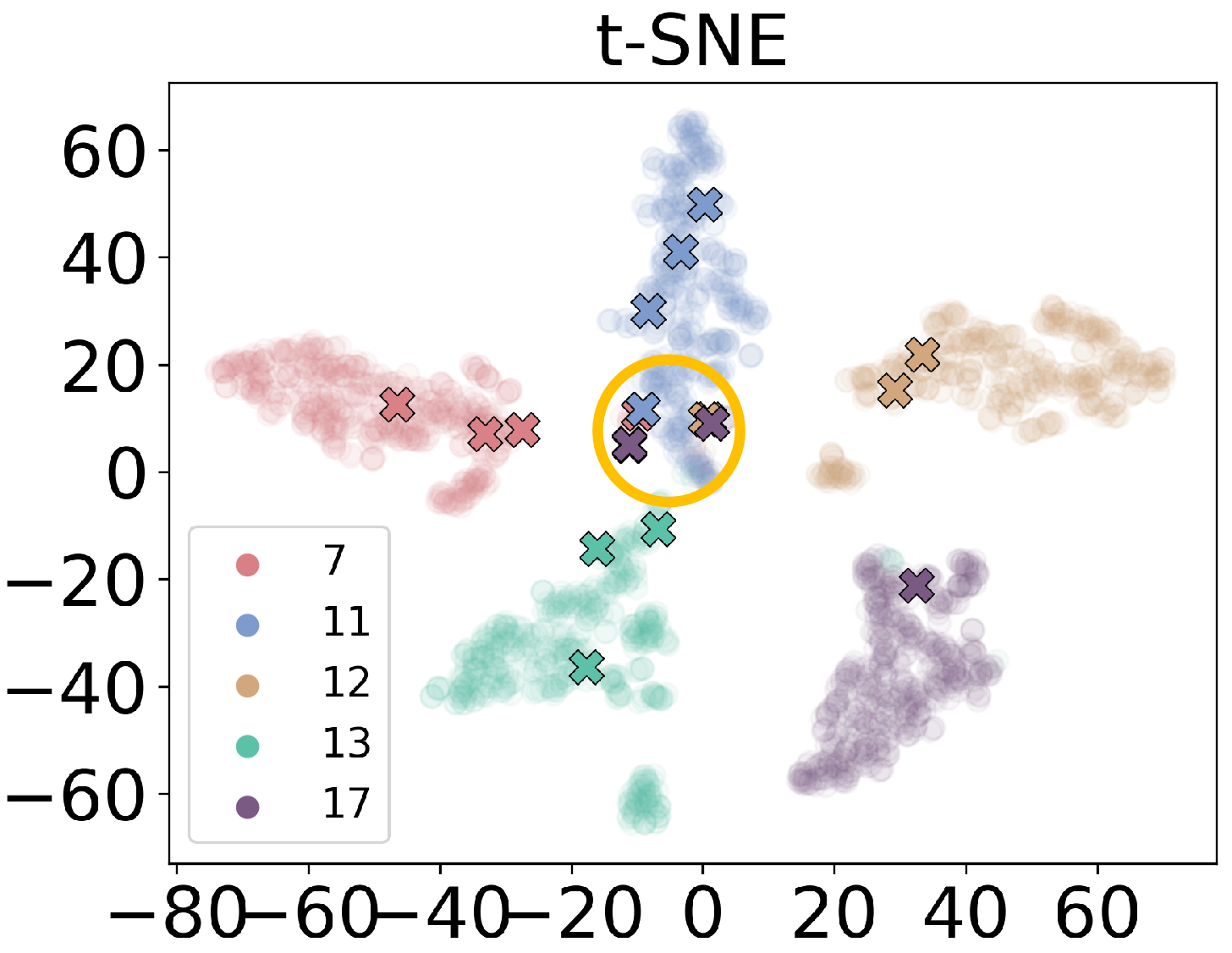}

		\end{minipage}
	}%
	\subfigure[Ours w/ BRPG]{
		\begin{minipage}[t]{0.49\linewidth}
		\centering
			\includegraphics[width=1\linewidth]{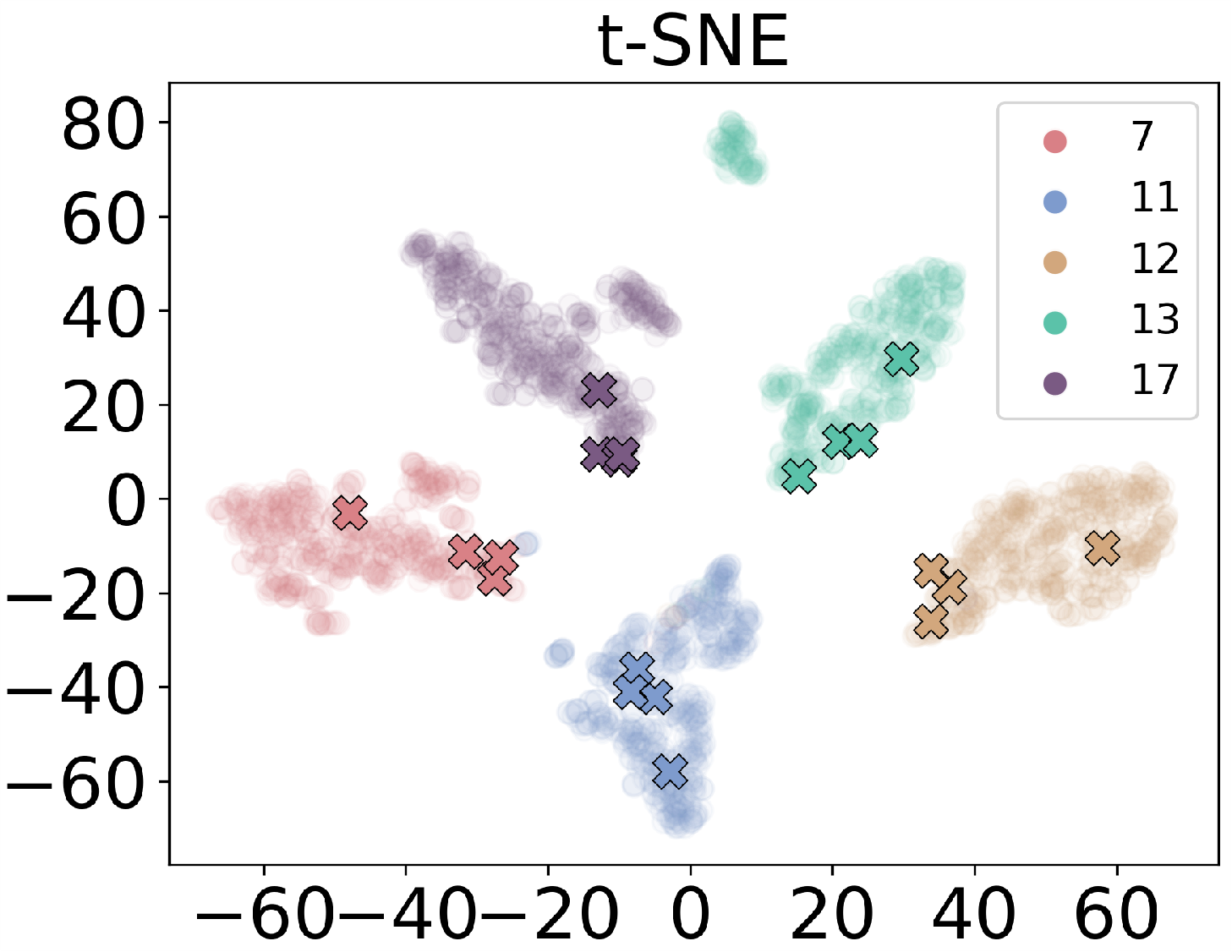}

		\end{minipage}
	}

	\caption{t-SNE \cite{tsne} visualization of feature distributions under the 1/4 (366) partition of PASCAL VOC 2012. Various models are trained using different settings: (a) SupOnly: Only supervised training. (b) $L_{unsup}$: Semi-supervised training incorporating $L_{unsup}$. (c) Plain $L_{pro}$: Prototype-based contrastive learning framework employing random sampling and direct K-Means clustering. (d) Ours w/ BRPG: Our training pipeline with BRPG. The associations between the displayed IDs and their respective semantic categories are: \{7: ``car", 11: “diningtable”, 12: “dog”, 13: “horse”, 17: “sheep”\}.}
	\label{fig5}
\end{figure}

\subsection{Qualitative Analysis}
In this section, qualitative analyses are performed to illustrate the impact of our approach in the feature space and visualize the segmentation performance, thereby further validating its effectiveness. All the experiments are conducted with DeepLabV3+.

\textbf{Distribution of feature representation.} The core idea of our method is to generate prototypes that refine classification boundaries in the feature space. Thus, it is crucial to explore the influence of our approach on feature distributions. Fig.~\ref{fig5} shows the feature distributions of various methods for multiple categories in PASCAL VOC 2012. It can be observed that the feature distributions of each class generated by the models of ``Plain $L_{pro}$" and ``Ours" are more compact than other methods. However, we observe that the prototype distributions of the two methods exhibit significant differences.

As shown in Fig.~\ref{fig5} (c), multiple prototypes from different categories are clustered within the small ``conflict" region (highlighted in the yellow circle), along with a number of challenging samples. This situation may confuse the network in feature classification. In contrast, our method exhibits fewer ``conflicts" in Fig.~\ref{fig5} (d), with the majority of prototypes located at the edges of feature distributions, indicating clear class boundaries. This corresponds to the superior performance of our model, as reported in the quantitative results. 

The different observations can be interpreted as follows: The prototypes initialized by ``Plain $L_{pro}$" tend to be closer to the semantic centers of the classes but further from the classification boundaries, which makes the network more vulnerable to the challenging samples and accordingly will negatively affect the feature generation and prototype update procedures. In contrast, our approach generates prototypes that are closer to the classification boundaries, which alleviates the problem of challenging samples and leads to a more stable training process.

\begin{figure*}[!t]
\centering
\includegraphics[width=0.63\paperwidth]{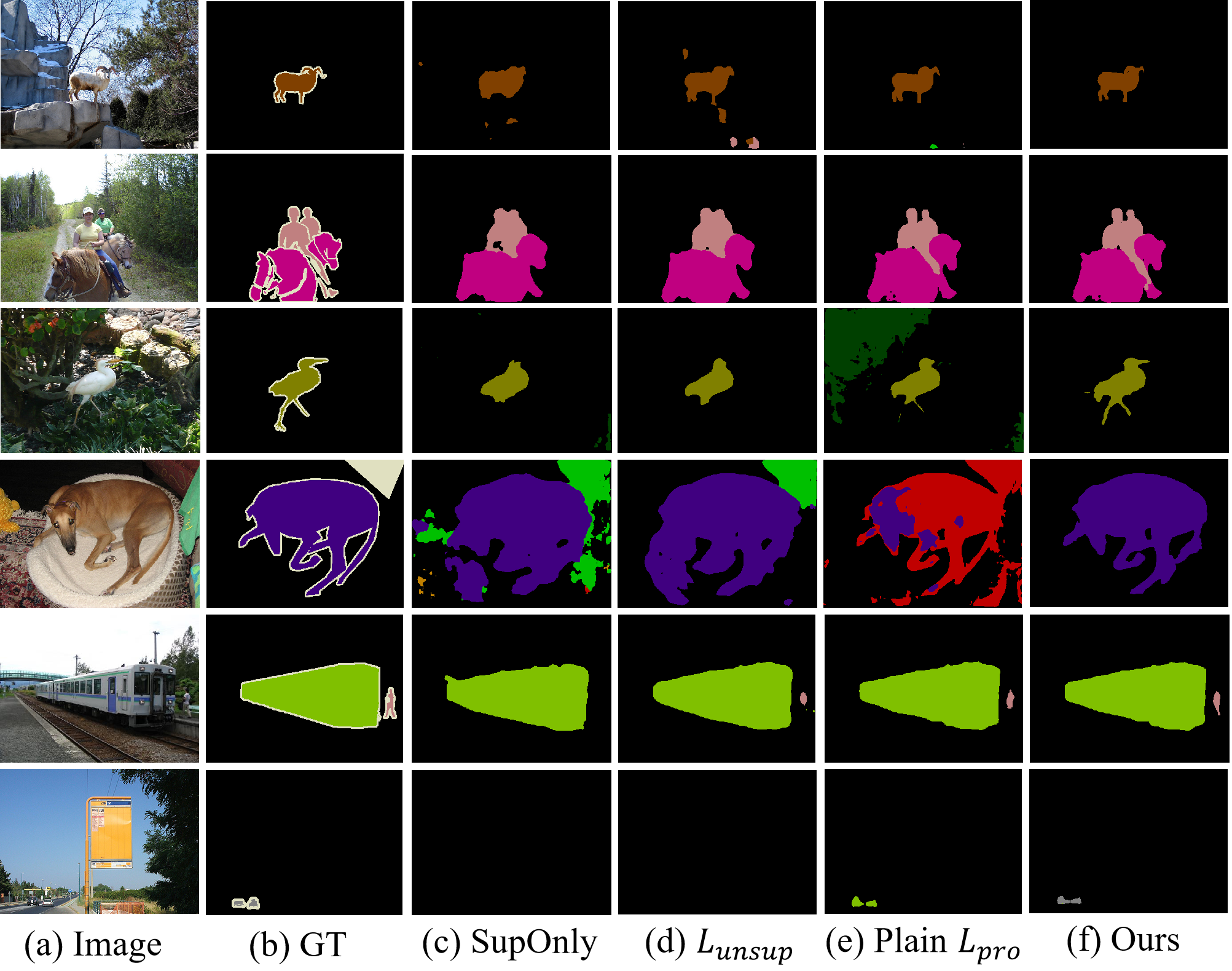}
\caption{Qualitative results on the \textbf{\textit{classic}} setting of \textbf{PASCAL VOC 2012} under the 1/4 partition protocol. (a) Input images. (b) Ground-truth. (c) Supervised-only model. (d) Semi-supervised model without prototype-based learning. (e)  Plain prototype-based model. (f) Our proposed model with BRPG. }
\label{fig6}
\end{figure*}

\begin{figure*}[!t]
\centering
\includegraphics[width=0.63\paperwidth]{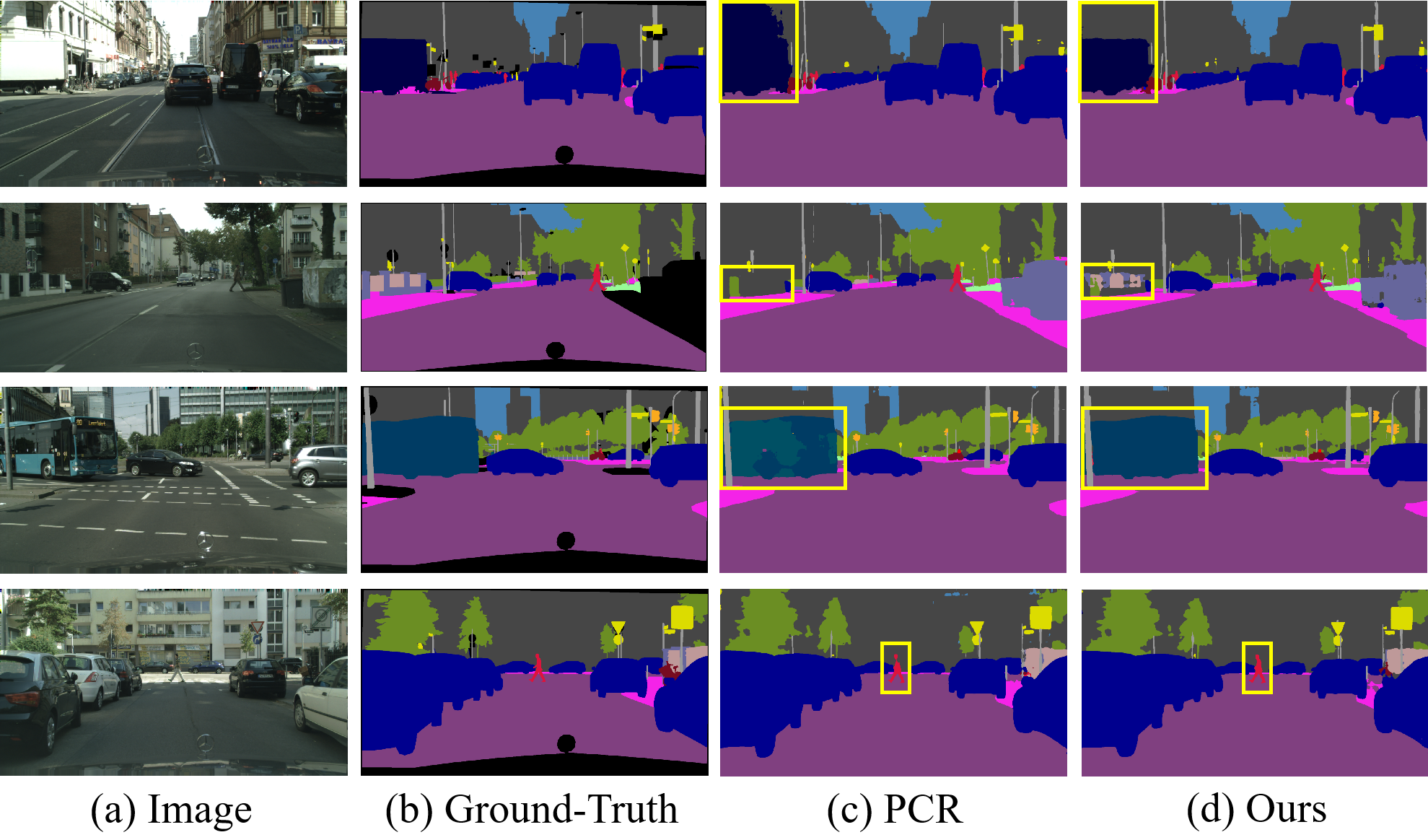}
\caption{Qualitative results on \textbf{Cityscapes} under the 1/16 partition protocol. (a) Input images. (b) Ground-truth. (c) the SOTA prototype-based method PCR. (d) Our method. Yellow rectangles highlight the improved segmentation results achieved by our approach.}
\label{fig7}
\end{figure*}

\textbf{Qualitative results on different components.} Fig.~\ref{fig6} presents the qualitative results when using different components of the proposed approach. It can be seen that the SupOnly model (c) is prone to yield incomplete and noisy predictions, whereas the semi-supervised model (d) produces more robust segments. The prototype-based method (e) effectively refines the object boundaries with a superior perception of class semantics. However, it may introduce noise (Row 3) and exhibit misclassifications (Row 4) in scenarios with cluttered backgrounds. 
In contrast, our method (f) generates finer object boundaries and demonstrates cleaner and more accurate predictions compared to (e). Furthermore, superior performance can be achieved in small object segmentation tasks (Row 5 and Row 6) with our approach. These visualization results further support the effectiveness of BRPG in perceiving category boundaries.

\textbf{Qualitative results on relevant methods.} In the qualitative comparison results on Cityscapes (Fig.~\ref{fig7}), our method outperforms the prototype-based method PCR \cite{pcr} in several aspects: (1) Enhanced accuracy in predicting challenging object boundaries. For instance, in the first row, our method successfully distinguishes between the white truck and the white building. (2) Increased capability to identify additional object classes, e.g., the fence and wall highlighted in the second row. (3) Improved semantic consistency within objects, shown in the third row where our approach achieves consistent category predictions within the bus, while the comparison method predicts different categories within the same object. (4) Superior edge details in object predictions, such as the person highlighted in the fourth row. These findings demonstrate the superiority of the proposed method as a prototype-based approach.

\begin{figure*}[!t]
\centering
\includegraphics[width=0.5\paperwidth]{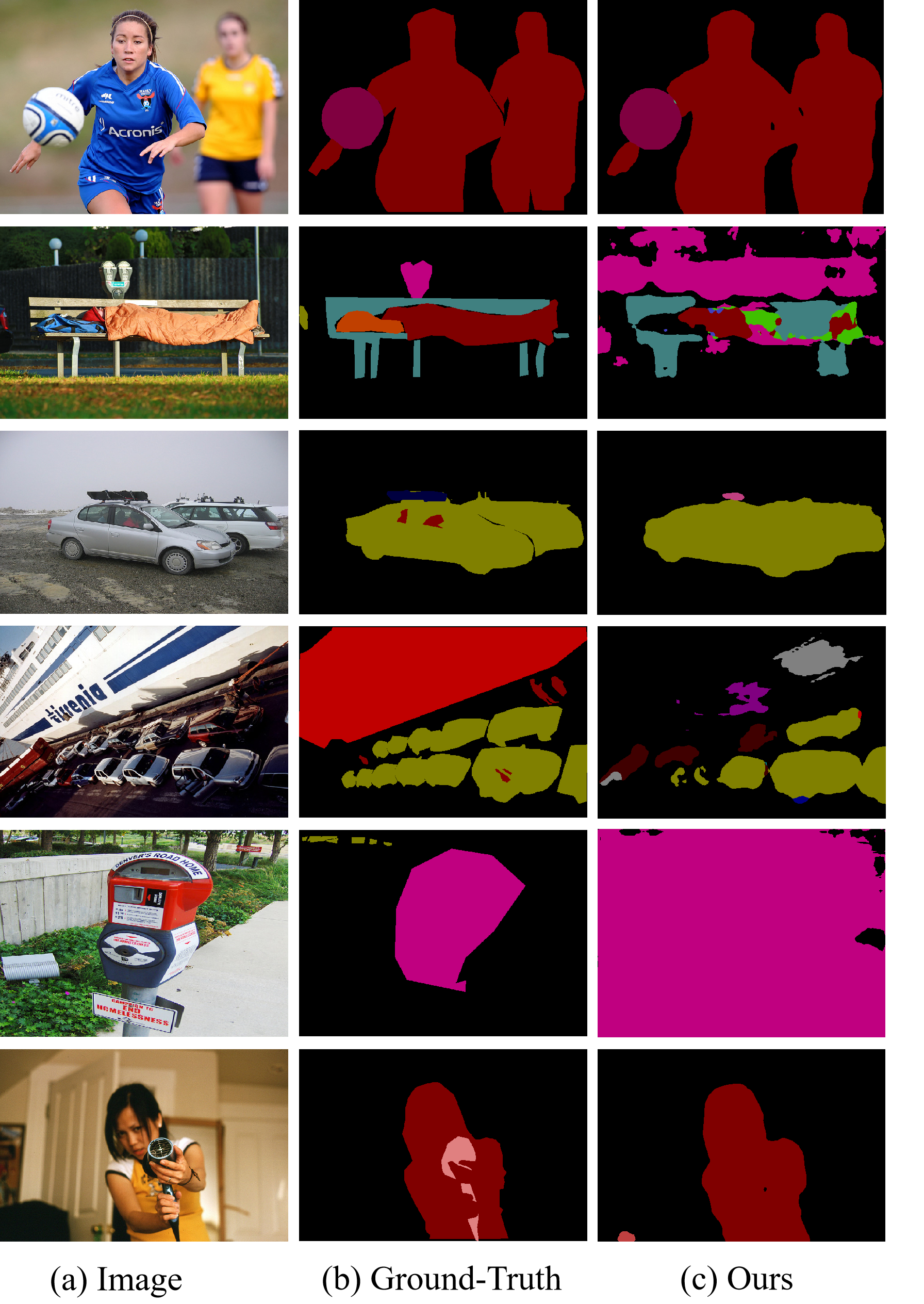}
\caption{Visualization of successful and failed cases on \textbf{MS COCO} under the 1/32 partition protocol. (a) Input images. (b) Ground-truth. (c) Segmentation results using our method. }
\label{fig-failure-case}
\end{figure*}

\textbf{Analysis of failure cases.} Our approach has been extended to the challenging large-scale MS COCO dataset, demonstrating impressive performance. However, the results presented in Table~\ref{tab:coco result} indicates the potential for improving the segmentation accuracy of our method. Fig.~\ref{fig-failure-case} illustrates the visualization results of successful and failed cases on the COCO dataset. The proposed method exhibits satisfactory segmentation  results on iconic-object images (objects centered in the image), as observed in the first row with people and the third row with cars. Nevertheless, in complex scenes with occlusion (Row 2, person covered by a blanket) and non-canonical viewpoints (Row 4), the objects cannot be accurately segmented. This limitation may be attributed to the lack of effective spatial augmentations in the model to adapt to diverse scenes. Furthermore, inferior performance can be observed in categories with limited training samples, even under canonical viewpoints. For instance, the failure to delineate the edges of the parking meter in Row 5 and to recognize a hairdryer in Row 6. This issue could stem from the generation of prototypes lacking class semantics in undertrained categories, and such prototypes may not effectively guide the model to make correct predictions. Hence, it is essential to explore class-specific prototype generation and training strategies to enhance the performance on underperforming categories.

\section{Conclusion}
In this paper, a novel end-to-end prototype-based approach for semi-supervised semantic segmentation is proposed, which improves the segmentation performance through boundary-refined prototype generation (BRPG). Specifically, we investigate the relationship between feature distributions and predicted confidence and propose the confidence-based prototype generation (CPG) method. CPG categorizes extracted features into high-confidence and low-confidence groups, followed by online sampling and clustering. In addition, an adaptive prototype optimization (APO) method is introduced to increase the number of cluster centers for categories with scattered feature distributions, further refining the class boundaries. The initialized prototypes are then utilized to guide the contrastive learning during model training. The experimental results demonstrate the effectiveness and generalization ability of our proposed method.

In future work, as discussed in Section \ref{sec:ablation study}, an optimal sampling strategy based on class-specific confidence thresholding will be further explored. Moreover, given the imbalanced class distributions in the datasets, more class-specific methods are worth studying.

\section*{Acknowledgments}
This work was supported by the Chinese National Natural Science Foundation under Grant 62076033 and U1931202.

\label{}

\appendix

\section{More Ablation Studies}\label{appendix: A more ablation study}

In this section, more ablation studies are conducted to provide the surplus process of parameter settings. All the experiments are conducted on 1/4 (366) split of PASCAL VOC 2012, employing DeepLabV3+ as the segmentation network.

\begin{table}[!ht]
\caption{Ablation study on the hyper-parameter $sample\_num$.}
    \label{tab:sample num}
    \centering
    \begin{tabular}{c|cccc}
    \toprule
        $sample\_num$ & 1000 & 3000 & 5000 & 10000 \\ \midrule
        
        VOC (366) & 77.39 & 77.30 & \textbf{78.00} & 77.77  \\ \bottomrule
    \end{tabular}
\end{table}

\begin{table}[!ht]
\caption{Ablation study on the storage capacity $V$ of memory banks.}
    \label{tab:memory bank capacity}
    \centering
    \begin{tabular}{c|cccc}
    \toprule
        $V$ & 10000 & 20000 & 30000 & 50000 \\ \midrule
        
        VOC (366) & 77.73 & 77.42 & \textbf{78.00} & 77.69  \\ \bottomrule
    \end{tabular}
\end{table}

\textbf{Configurations for feature update.} The hyper-parameter $sample\_num$, as detailed in Section \ref{sec:CPG}, regulates the update speed of stored features, while the storage capacity $V$ of memory banks determines their preservation range. Table~\ref{tab:sample num} and Table~\ref{tab:memory bank capacity} present the impacts of these two parameters on model performance. As observed, the best segmentation results can be obtained with the settings of $sample\_num=5000$ and $V=30000$.

\begin{table}[!ht]
\caption{Ablation study on the value of confidence threshold $\eta_t$. In the linear strategy, the values of $\eta_0$ and $\eta_e$ are denoted by [${\eta_0}, {\eta_e}$].}
    \label{tab:confidence threshold eta_t}
    \centering
    \begin{tabular}{c|ccccc}
    \toprule
        $\eta_t$ & 0.7 & 0.75 & 0.8 & 0.85 & 0.9 
        
        \\ \midrule
        
        VOC (366) & 77.06 & 77.03 & 77.50 & 77.69 & 77.61

        \\ \midrule 
        \addlinespace[0.1ex]
        \midrule

        $\eta_t$ & 0.95 & $\left[0.8, 0.9\right]$ & $\left[0.8, 0.95\right]$ & $\left[0.85, 0.95\right]$ & $\left[0.9, 0.95\right]$ 
        \\ \midrule
        VOC (366) & 77.63 & 77.79 & \textbf{78.00} & 77.58 & 77.57
        \\ \bottomrule
        
    \end{tabular}
\end{table}

\begin{table}[!ht]
\caption{Ablation study on different numbers of added prototypes.}
    \label{tab:proto add num}
    \centering
    \begin{tabular}{c|ccccc}
    \toprule
        $n_{add}$ & 0 & 1 & 2 & 3 & 4\\ \midrule
        
        VOC (366) & 77.39 & \textbf{78.00} & 77.52 & 77.40 & 77.72 \\ \bottomrule
    \end{tabular}
\end{table}

\textbf{Value of the confidence threshold $\eta_t$.} In Section \ref{sec:Training with Prototype-Based Contrastive Learning}, a linear strategy denoted by $\eta_{t}=\eta_{0}+\left(\eta_{e}-\eta_{0}\right) \frac{{ curr\_iter }}{ { total\_iter }}$ is employed to adjust the confidence threshold for samples involved in the calculation of prototype-based loss. To investigate the influence of $\eta_t$ on segmentation results, we conduct ablation studies with two settings: fixed value and linear strategy, as shown in Table~\ref{tab:confidence threshold eta_t}. Under the fixed value setting, superior performance is observed when $\eta_t\geq0.8$. This may be attributed to the mitigation of overfitting to the noise in pseudo-labels. Additionally, the linear strategy consistently yields satisfactory results when $\eta_0\geq0.8$. In our experiments, we adopt $\eta_0=0.8$ and $\eta_e=0.95$ as the default setting to achieve optimal performance.

\textbf{Another form of APO.} APO aims to increase the number of prototypes for categories with scattered feature distributions, as denoted by $N_{\alpha, c}=n_{0 \alpha}+\mathbbm{1}\left(D_{\operatorname{sim}, \alpha, c}<\gamma_{t}\right)$ in Eq.~(\ref{eq:N erfa c sim}). However, it can also be modified to $N_{\alpha, c}=n_{0 \alpha}+n_{add} \cdot \left(D_{\operatorname{sim}, \alpha, c}<\gamma_{t}\right)$ with a hyper-parameter $n_{add}$ to investigate the effect of varying the number of added prototypes. As shown in Table~\ref{tab:proto add num}, APO exhibits its optimal effectiveness when $n_{add}$ is set to 1.

\bibliographystyle{elsarticle-num-names} 
\bibliography{ref}

\end{document}